%% file: icml2023.tex
\documentclass{article}

\input{math_commands.tex}

\usepackage{hyperref}
\usepackage{url}
\usepackage{amsmath}
\usepackage{amsthm}
\usepackage{microtype}
\usepackage{graphicx}
\usepackage{mathtools}
\usepackage{amsthm}
\usepackage{bibentry}
\usepackage{array}
\usepackage{verbatim}
\usepackage{enumitem}
\usepackage{amssymb}
\usepackage{subcaption}
\usepackage{booktabs} % for professional tables
\usepackage{mathtools}
\usepackage{wrapfig}
\usepackage{placeins}
\usepackage[nolist,nohyperlinks]{acronym}
\usepackage[accepted]{icml2023}

\usepackage[capitalize,noabbrev]{cleveref}

\theoremstyle{plain}

\theoremstyle{definition}

\theoremstyle{remark}

\newcommand{\highlight}[1]{\colorbox{blue!10}{#1}}

\definecolor{mygray}{gray}{0.4}

\newcommand{\g}[2]{#1\textsubscript{\textcolor{mygray}{$\pm$#2}}}
\acrodef{GFlowNets}[GFlowNets]{\emph{Generative Flow Networks}}
\acrodef{MOGFN}[MOGFN]{\emph{Multi-Objective GFlowNets}}
\acrodef{MOO}[MOO]{\emph{Multi-Objective Optimization}}

\usepackage[textsize=tiny]{todonotes}

\icmltitlerunning{Multi-Objective GFlowNets}

\begin{document}

\twocolumn[
\icmltitle{Multi-Objective GFlowNets}



\icmlsetsymbol{note}{*}

\begin{icmlauthorlist}
\icmlauthor{Moksh Jain}{udem,mila}
\icmlauthor{Sharath Chandra Raparthy}{udem,mila,note}
\icmlauthor{Alex Hernandez-Garcia}{udem,mila}
\icmlauthor{Jarrid Rector-Brooks}{udem,mila}
\icmlauthor{Yoshua Bengio}{udem,mila,cifar}
\icmlauthor{Santiago Miret}{intel}
\icmlauthor{Emmanuel Bengio}{recursion}
\end{icmlauthorlist}

\icmlaffiliation{udem}{Universit\'e de Montr\'eal}
\icmlaffiliation{mila}{Mila - Quebec AI Institute}
\icmlaffiliation{recursion}{Recursion}
\icmlaffiliation{intel}{Intel Labs}
\icmlaffiliation{cifar}{CIFAR Fellow \& IVADO}

\icmlcorrespondingauthor{Moksh Jain}{mokshjn00@gmail.com}

\icmlkeywords{generative flow networks, multi-objective optimization, scientific discovery}

\vskip 0.3in
]



\printAffiliationsAndNotice{$^*$Work done during an internship at Recursion}  % leave blank if no need to mention equal contribution

\begin{abstract}
We study the problem of generating \emph{diverse} candidates in the context of Multi-Objective Optimization. In many applications of machine learning such as drug discovery and material design, the goal is to generate candidates which simultaneously optimize a set of potentially conflicting objectives. Moreover, these objectives are often imperfect evaluations of some underlying property of interest, making it important to generate diverse candidates to have multiple options for expensive downstream evaluations. We propose Multi-Objective GFlowNets (MOGFNs), a novel method for generating diverse Pareto optimal solutions, based on GFlowNets. 
We introduce two variants of MOGFNs: MOGFN-PC, which models a family of independent sub-problems defined by a scalarization function, with reward-conditional GFlowNets, and MOGFN-AL, which solves a sequence of sub-problems defined by an acquisition function in an active learning loop. Our experiments on wide variety of synthetic and benchmark tasks demonstrate advantages of the proposed methods in terms of the Pareto performance and importantly, improved candidate diversity, which is the main contribution of this work.
\end{abstract}

\input{tex/intro_iclr}

\input{tex/background}

\input{tex/mogfn}

\input{tex/related}

\input{tex/empirical_results}
\subsection{Synthetic Tasks}

\input{tex/grid}

\input{tex/string}

\subsection{Benchmark Tasks}

\input{tex/mol_exp}
\input{tex/dna}

\input{tex/al}

\input{tex/analysis}

\section{Conclusion}
In this work, we study Multi-Objective GFlowNets (MOGFN) for the generation of diverse Pareto-optimal candidates. We present two instantiations of MOGFNs: MOGFN-PC, which models a family of independent single-objective sub-problems with conditional GFlowNets, and MOGFN-AL, which optimizes a sequence of single-objective sub-problems. We demonstrate the efficacy of MOGFNs empirically on wide range of tasks ranging from molecule generation to protein design.

% The analysis in \Cref{sec:analysis} hints that the distribution of sampling preferences $p(\omega)$ affects the Pareto performance. Although uniform coverage over the space of preferences gives promising performance, in real world applications we might be interested in only specific region of the Pareto front. An interesting future work could be to explore gradient based techniques for learning preferences for more structured exploration in preference space.  Moreover, for our active learning experiments future work should focus on the development of preference-conditioned acquisiton functions. 
As a limitation, we note that the benefits of MOGFNs are limited in problems where the rewards have a single mode (see \cref{sec:dna-exps}). 
% The analysis in \Cref{sec:analysis} hints that the distribution of sampling preferences $p(\omega)$ affects the Pareto performance. 
For certain practical applications, we are interested only in a specific region of the Pareto front. Future work may explore gradient-based techniques to learn $p(\omega)$ for more structured exploration of the preference space. Within the context of MOFGN-AL, an interesting research avenue is the development of preference-conditional acquisition functions. 

\paragraph{Code} The code for the experiments is available at \url{https://github.com/GFNOrg/multi-objective-gfn}.

\section*{Acknowledgements}
We would like to thank Nikolay Malkin, Yiding Jiang and Julien Roy for valuable feedback and comments. The research was enabled in part by computational resources provided by the Digital Research Alliance of Canada (\url{https://alliancecan.ca/en}) and Mila (\url{https://mila.quebec}). We also acknowledge funding from CIFAR, IVADO, Intel, Samsung, IBM, Genentech, Microsoft. 

\bibliography{icml2023}
\bibliographystyle{icml2023}

%%%%%%%%%%%%%%%%%%%%%%%%%%%%%%%%%%%%%%%%%%%%%%%%%%%%%%%%%%%%%%%%%%%%%%%%%%%%%%%
%%%%%%%%%%%%%%%%%%%%%%%%%%%%%%%%%%%%%%%%%%%%%%%%%%%%%%%%%%%%%%%%%%%%%%%%%%%%%%%
% APPENDIX
%%%%%%%%%%%%%%%%%%%%%%%%%%%%%%%%%%%%%%%%%%%%%%%%%%%%%%%%%%%%%%%%%%%%%%%%%%%%%%%
%%%%%%%%%%%%%%%%%%%%%%%%%%%%%%%%%%%%%%%%%%%%%%%%%%%%%%%%%%%%%%%%%%%%%%%%%%%%%%%
\newpage
\appendix
\onecolumn
\input{tex/appendix}

\end{document}

%% file: math_commands.tex
\usepackage{amsmath,amsfonts,bm}

\newcommand{\Tau}[0]{{\mathcal{T}}}

\def\eqref#1{equation~\ref{#1}}
\def\1{\bm{1}}

\DeclareMathAlphabet{\mathsfit}{\encodingdefault}{\sfdefault}{m}{sl}
\SetMathAlphabet{\mathsfit}{bold}{\encodingdefault}{\sfdefault}{bx}{n}

%% file: tex/intro_iclr.tex
\section{Introduction}
\label{sec:intro}
% \emph{Multi-Objective Optimization}~\citep[MOO;][]{ehrgott2005multicriteria,miettinen2012nonlinear} arises in multiple problems of practical relevance.
Decision making in practical applications usually involves reasoning about multiple, often conflicting, objectives~\citep{keeney1993decisions}. Consider the example of in-silico drug discovery, where the goal is to generate \emph{novel} drug-like molecules that effectively inhibit a target, are easy to synthesize, and possess a safety profile for human use~\citep{dara2021machine}. These objectives often exhibit mutual incompatibility as molecules that are effective against a target may also have detrimental effects on humans, making it infeasible to find a single molecule that maximizes all the objectives simultaneously. Instead, the goal in these Multi-Objective Optimization~\citep[MOO;][]{ehrgott2005multicriteria,miettinen2012nonlinear} problems is to identify candidate molecules that are Pareto optimal, covering the best possible trade-offs between the objectives. 
% Such problems fall under the umbrella of \emph{Multi-Objective Optimization}~\citep[MOO;][]{ehrgott2005multicriteria,miettinen2012nonlinear}. %These objectives also merely serve as imperfect proxies for the underlying 
%Such problems fall under the umbrella of \emph{Multi-Objective Optimization}~\citep[MOO;][]{ehrgott2005multicriteria,miettinen2012nonlinear}, wherein 
% In such cases, one is interested in identifying \emph{Pareto-optimal} candidates. The set of Pareto-optimal candidates covers all the best tradeoffs among the objectives, known as the Pareto front, where each point on that front corresponds to a different set of weights associated with each of the objectives.

A less appreciated aspect of multi-objective problems is that the objectives to optimize are usually underspecified proxies which only approximate the true design objectives. For instance, the binding affinity of a molecule to a target is an imperfect approximation of the molecule's inhibitory effect against the target in the human body.%, and therefore include intrinsic epistemic uncertainty associated with their predictions.
{} In such scenarios it is important to not only cover the Pareto front but also to generate sets of \emph{diverse} candidates for each Pareto optimal solution 
% corresponding to each trade-off on the Pareto front 
in order to increase the likelihood of success of the generated candidates in expensive downstream evaluations, such as in-vivo tests and clinical trials~\citep{jain2022biological}. 
% This is important because the objectives are imperfect. if different kinds of candidates were to satisfy such proxy objectives (for one point on the Pareto front, one trade-off between the different measured objectives) we should try to obtain samples from all these kinds of candidates. This would avoid putting all our eggs in the same basket and end up with having chosen the wrong kind (which fails on downstream experiments).
% As the objectives are imperfect, it is important that we obtain samples from a diverse set of candidates. This is necessary as multiple sets of candidates may satisfy the proxy objectives for a given point on the Pareto front. 

The benefits of generating diverse candidates are twofold. First, by diversifying the set of candidates % we increase the chances of success in downstream experiments and reduce the risk of failure due to selecting an unsuitable candidate %.
%
%%%% <Emmanuel> I'm not sure I'm expressing this correctly. I think it's important to say that diversity is wise because at a given point x the proxy is making some amount of error, and points near x are likely to have the same "generalization errors" as x. In that sense, if we sample far from x, at least we'll be betting on different errors that the model is making -- and the model is bound to get it right sometimes!
we obtain an advantage similar to Bayesian ensembles: we reduce the risk of failure that might occur due to the imperfect generalization of learned proxy models. Diverse candidates should lie in different regions of the input-space manifold where the objective of interest might be large (considering the uncertainty in the output of the proxy model). 
% (where the models might make different, perhaps less wrong assumptions). 
Second, experimental measurements such as in-vitro assays may not reflect the ultimate objectives of interest, such as efficacy in human bodies. Multiple candidates may have the same assay score, but different downstream efficacy, so diversity in candidates increases odds of success. Existing approaches for MOO overlook this aspect of diversity and instead focus primarily on generating Pareto optimal solutions.
 
%\rebuttal{In the context of MOO, this amounts to finding all the candidates corresponding to each point on the Pareto front.}

Generative Flow Networks~\citep[GFlowNets;][]{bengio2021flow,bengio2021gflownet} are a recently proposed family of probabilistic models which tackle the problem of diverse candidate generation. Contrary to the reward maximization view of prevalent Reinforcement Learning (RL) and Bayesian optimization (BO) approaches, GFlowNets sample candidates with probability proportional to their reward. Sampling candidates, as opposed to greedily generating them implicitly encourages diversity in the generated candidates. GFlowNets have shown promising results in single objective problems such as molecule generation~\citep{bengio2021flow} and biological sequence design~\citep{jain2022biological}.

% In this paper, we study \emph{Multi-Objective GFlowNets} (MOGFNs), extensions of GFlowNets which tackle the multi-objective optimization problem. We thus contribute to both extending MOO to obtain a diversity of solutions for each point on the Pareto front and extending GFlowNets to explore and sample from the Pareto front. 
In this paper, we propose Multi-Objective GFlowNets (MOGFNs), which leverage the strengths of GFlowNets and existing MOO approaches to enable the generation of diverse Pareto optimal candidates. We consider two variants of MOGFNs -- (a) \emph{Preference-Conditional GFlowNets} (MOGFN-PC) which leverage the decomposition of MOO into single objective sub-problems through scalarization, and (b) MOGFN-AL, which leverages the transformation of MOO into a sequence of single objective sub-problems within the framework of multi-objective Bayesian optimization. 
% We empirically demonstrate the advantage of MOGFNs over existing approaches on a variety of high-dimensional multi-objective optimization tasks: the generation of small molecules, DNA aptamer sequences and fluorescent proteins. 
Our contributions are as follows:

\vspace{-2mm}
\begin{enumerate}[leftmargin=6mm]
\setlength\itemsep{0.1pt}
    \item [\textbf{C1}] We introduce a novel framework of MOGFNs to tackle the practically significant and previously unexplored problem of \emph{diverse} candidate generation in MOO.
    % We demonstrate how two variants of MOGFNs  -- MOGFN-PC and MOGFN-AL -- can be succesfully applied to Multi-Objective Optimization. 
    % \item [\textbf{C2}]  We empirically demonstrate the efficacy of MOGFN-PC in high-dimensional MOO problems. Our work is the first successful empirical validation of Reward-Conditional GFlowNets~\citep{bengio2021gflownet}.
    \item [\textbf{C2}] Through experiments on challenging molecule generation and sequence generation tasks, we demonstrate that MOGFN-PC generates \emph{diverse} Pareto-optimal candidates. This is the first successful application and empirical validation of reward-conditional GFlowNets~\citep{bengio2021gflownet}.
    
    \item  [\textbf{C3}] In a challenging active learning task for designing fluorescent proteins, we show that MOGFN-AL results in significant improvements to sample efficiency as well as diversity of generated candidates. % over baselines. %for generating \emph{diverse} Pareto-optimal candidates. %Our work is the first to leverage \ac{GFlowNets} for mutation-based biological sequence design.
    \item [\textbf{C4}] We perform a thorough analysis on the key components of MOGFNs to provide insights into design choices that affect performance. 
\end{enumerate}

%% file: tex/background.tex
% Note to co-authors: in the GFlowNet Foundations paper, | is used instead of \mid. I suggest to always use | for consistency, as \mid inserts additional horizontal spaces.
\section{Background}
% This section presents the essential background on multi-objective optimization and GFlowNets before presenting multi-objective GFlowNets in Section~\ref{sec:mogfn}.
\label{sec:background}
\subsection{Multi-Objective Optimization}
\label{sec:moo_bg}
Multi-objective Optimization (MOO) involves finding a set of feasible candidates $x^\star \in \mathcal{X}$ which simultaneously maximize $d$ objectives $\mathbf{R}(x) = [R_1(x), \dots, R_d(x)]$: 
\begin{equation}
    \max_{x \in \mathcal{X}} \mathbf{R}(x).
    \label{eq:moo}
\end{equation}
When these objectives are conflicting, there is no single $x^\star$ which simultaneously maximizes all objectives. 
Consequently, MOO adopts the concept of \emph{Pareto optimality}, which describes a \emph{set of solutions} that provide optimal trade-offs among the objectives. 

Given $x_1, x_2 \in \mathcal{X}$, $x_1$ is said to \emph{dominate} $x_2$, written ($x_1 \succ x_2$), \emph{iff} $R_i(x_1) \ge R_i(x_2) \enspace \forall i \in \{1, \dots, d\}$ and $\exists k \in \{1, \dots, d\}$ such that $R_k(x_1) > R_k(x_2)$. 
A candidate $x^\star$ is \emph{Pareto-optimal} if there exists no other solution $ x' \in \mathcal{X}$ which dominates $x^\star$. In other words, for a Pareto-optimal candidate it is impossible to improve one objective without sacrificing another. 
The \emph{Pareto set} is the set of all Pareto-optimal candidates in $\mathcal{X}$, and the \emph{Pareto front} is defined as the image of the Pareto set in objective-space.

\paragraph{Diversity:} Since the objectives are not guaranteed to be \emph{injective}, any point on the Pareto front can be the image of several candidates in the Pareto set. This designates 
% is one natural 
\emph{diversity in candidate space}. Capturing all the diverse candidates, corresponding to a point on the Pareto front, is critical for applications such as in-silico drug discovery, where the objectives $R_i$ (e.g. binding affinity to a target protein) are mere proxies for the more expensive downstream measurements (e.g., effectiveness in clinical trials on humans). This notion of diversity of candidates is typically not captured by existing approaches for MOO. 

% In practice, MOO problems can be challenging when they have high-dimensional objectives and candidates. 
\subsubsection{Approaches for tackling MOO}
While, there exist  many approaches tackling MOO problems~\citep{ehrgott2005multicriteria,miettinen2012nonlinear,pardalos2017non}, in this work, we consider two distinct approaches that decompose the MOO problem into a family of single objective sub-problems. These approaches, described below, are well-suited for the GFlowNet formulations we introduce in \Cref{sec:mogfn}.

\paragraph{Scalarization:} In scalarization, a set of weights (preferences) $\omega_i$ are assigned to the objectives $R_i$, where $\omega_i \ge 0$ and $\sum_{i=1}^d\omega_i =1$. 
%Any other MOO approach that also quantifies each point on the Pareto front could be used with the methods we propose below. With scalarization, t
The MOO problem can then be decomposed into single-objective sub-problems of the form $\max_{x \in \mathcal{X}}  R(x|\omega)$, where $R(x|\omega)$ is called a \emph{scalarization function}, which combines the $d$ objectives into a scalar. Solutions to these sub-problems capture all Pareto-optimal solutions to the original MOO problem depending on the choice of $R(x|\omega)$ and characteristics of the Pareto front. \emph{Weighted Sum Scalarization}, $R(x|\omega) =\sum_{i=1}^d\omega_iR_i(x)$, for instance, captures all Pareto optimal candidates for problems with a convex Pareto front~\citep{ehrgott2005multicriteria}. On the other hand, \emph{Weighted Tchebycheff}, $R(x|\omega) = \max\limits_{1\le i\le d} {\omega_i | R_i(x) - z_i^\star|}$, where $z_i^\star$ denotes an ideal value for objective $R_i$, captures all Pareto optimal solutions even for problems with a non-convex Pareto front~\citep{choo1983proper,pardalos2017non}. 
% See \Cref{app:wls_scalarization} for more discussion on scalarization.  
As such, scalarization transforms the multi-objective optimization into a family of \emph{independent} single-objective sub-problems.

\paragraph{Multi-Objective Bayesian Optimization:} In many applications, the objectives $R_i$ can be expensive to evaluate, thereby making sample efficiency essential. Multi-Objective Bayesian optimization (MOBO)~\citep{shah2016pareto,garnett_bayesoptbook_2022} builds upon BO to tackle these scenarios. MOBO relies on a probabilistic model $\hat{f}$ which approximates the objectives $\mathbf{R}$ (oracles). $\hat{f}$ is typically a multi-task Gaussian Process~\citep{shah2016pareto}. Notably, as the model is Bayesian, it captures the epistemic uncertainty in the predictions due to limited data available for training, which can be used as a signal for prioritizing potentially useful candidates. The optimization is performed over $M$ rounds, where each round $i$ consists of fitting the surrogate model $\hat{f}$ on the data $\mathcal{D}_i$ accumulated from previous rounds, and using this model to generate a batch of $b$ candidates $\{x_1, \dots,x_b\}$ to be evaluated with the oracles $\mathbf{R}$, resulting in $\mathcal{B}_i=\{(x_1, y_1), \dots, (x_b, y_b)\}$. The evaluated batch $\mathcal{B}$ is then incorporated into the data for the next round $\mathcal{D}_{i+1} = \mathcal{D}_i\cup \mathcal{B}$. The batch of candidates in each round is generated by maximizing an \emph{acquisition function} $a$ which combines the predictions from the surrogate model along with its epistemic uncertainty into a single scalar utility score. The acquisition function quantifies the utility of a candidate given the candidates evaluated so far. 
% Consider, for example, the Noisy Expected Hypervolume Improvement~\citep[NEHVI;][]{daulton2020differentiable} that quantifies the improvement in hypervolume bounded by the Pareto front that is expected under the model $\hat{f}$ upon acqusition of a candidate $x$.
Effectively, MOBO decomposes the MOO into a \emph{sequence} of single objective optimization problems of the form $\max_{\{x_1, \dots, x_b\}\in 2^\mathcal{X}}a(\{x_1, \dots, x_b\};\hat{f})$.

\subsection{GFlowNets}
\label{sec:gfn_bg}
GFlowNets~\citep{bengio2021flow,bengio2021gflownet} are a family of probabilistic models that learn a stochastic policy to generate \emph{compositional} objects $x \in \mathcal{X}$, such as a graph describing a candidate molecule, through a sequence of steps, with probability proportional to their reward $R(x)$. 
If $R:\mathcal{X}\to \mathbb{R}^+$ has multiple modes then i.i.d samples from $\pi \propto R$ gives a good coverage of the modes of $R$, resulting in a diverse set of candidate solutions.
The sequential construction of $x \in \mathcal{X}$ can be described as a trajectory $\tau\in \Tau$ in a weighted directed acyclic graph (DAG) $\mathcal{G}=\left(\mathcal{S}, \mathcal{E}\right)$\footnote{If the object is constructed in a canonical order (say a string constructed from left to right), $\mathcal{G}$ is a tree.}, starting from an empty object $s_0$ and following actions $a \in \mathcal{A}$ as building blocks. For example, a molecular graph may be sequentially constructed by adding and connecting new nodes or edges to the graph.
Let $s \in \mathcal{S}$, or state, denote a partially constructed object. Transitions
between states $s\xrightarrow[]{a} s' \in \mathcal{E}$ indicate that action $a$ at state $s$ leads to state $s'$. Sequences of such transitions form constructive trajectories. 

The GFlowNet forward policy $P_F(- | s)$ is a distribution over the children of state $s$. An object $x$ can be generated by starting at $s_0$ and sampling a sequence of actions iteratively from $P_F$. % is defined as $P_F(s'|s) = \frac{F(s\rightarrow s')}{F(s)}$. 
Similarly, the backward policy $P_B(- | s)$ is a distribution over the parents of state $s$ and can generate backward trajectories starting at any terminal $x$, e.g., iteratively sampling from $P_B$ starting at $x$ shows a way $x$ could have been constructed. %is given by $P_B(s|s') = \frac{F(s\rightarrow s')}{F(s')}$. 
Let $\pi(x)$ be the marginal likelihood of sampling trajectories terminating in $x$ following $P_F$, and partition function $Z=\sum_{x\in\mathcal{X}}R(x)$. % $\pi(x) = \frac{\sum_{x\rightarrow s_f \in \tau}F(\tau)}{Z}$
The learning problem solved by GFlowNets is to learn a forward policy $P_F$ such that the marginal likelihood of sampling any object $\pi(x)$ is proportional to its reward $R(x)$. 
% \cite{bengio2021flow} show that for a consistent flow $F$ with the terminal flows set equal to the rewards of the terminal node, $F(x\rightarrow s_f) = R(x)$, the forward policy generates $x$ with a probability proportional to it's reward $R(x)$, $\pi(x) \propto R(x)$. Consequently, GFlowNets learn a flow function that is \emph{consistent} where the terminal flows are equal to the rewards.
In this paper we adopt the \emph{trajectory balance}~\citep[TB;][]{malkin2022trajectory} learning objective. The trajectory balance objective learns $P_F(- | s;\theta), P_B(- | s;\theta), Z_\theta$ parameterized by $\theta$, which approximate the forward and backward policies and partition function such that $\pi(x)\approx \frac{R(x)}{Z}, \enspace \forall x\in \mathcal{X}$. We refer the reader to~\citet{bengio2021gflownet,malkin2022trajectory} for a more thorough introduction to GFlowNets. 
%A GFlowNet consists of the tuple $(P_F, P_B, Z)$, which approximate the forward and backward policies and partition function induced by a \emph{consistent flow}. Let $\theta$ denote the learnable parameters of the GFlowNet. $\theta$ can be learned with the \emph{trajectory balance} objective~\citep{malkin2022trajectory} 
% \begin{equation}
%     \mathcal{L}(\tau;\theta) = \left(\log \frac{Z_\theta\prod_{s\rightarrow s' \in \tau}P_F(s'|s;\theta)}{R(x)\prod_{s\rightarrow s' \in \tau}P_B(s|s';\theta)} \right)^2.
%     \label{eq:tb_loss}
% \end{equation}

% During training, trajectories are sampled from a mixture of the current forward policy $P_F$ and a uniform random policy $U$, with a parameter $\delta$ which controls the fraction of actions sampled from $U$. In many problem settings, one would like to focus on the modes, which can be achieved by exponentiation of the reward with a constant $\beta$, resulting in $\pi(x) \propto R(x)^{\beta}$.

%% file: tex/mogfn.tex
\section{Multi-Objective GFlowNets}
\label{sec:mogfn}

In this section, we introduce Multi-Objective GFlowNets (MOGFNs) to tackle the problem of \emph{diverse} Pareto optimal candidate generation. The two MOGFN variants we discuss respectively exploit the decomposition of MOO problems into a family of independent single objective subproblems or a sequence of single objective sub-problems. 

% We broadly categorize \emph{Multi-Objective GFlowNets} (MOGFNs) as GFlowNets which solve a family of sub-problems derived from a \ac{MOO} problem. We first consider solving a family of MOO sub-problems simultaneously with preference-conditional GFlowNets, followed by MOGFN-AL, which solves a sequence of MOO sub-problems.
% consider two instantiations: (a) Preference-Conditional GFlowNets (MOGFN-PC) which leverage weighted-sum scalarization and (b) MOGFN-AL which leverages multi-objective acquisition functions for active learning. 

\subsection{Preference-Conditional GFlowNets}
\label{sec:mogfn-pc}
As discussed in Section~\ref{sec:moo_bg}, given an appropriate scalarization function, candidates optimizing each sub-problem $\max_{x\in\mathcal{X}}R(x|\omega)$ correspond to a single point on the Pareto front. As the objectives are often imperfect proxies for some underlying property of interest we aim to generate \emph{diverse} candidates for each sub-problem. One naive way to achieve this is by solving each independent sub-problem with a separate GFlowNet. However, this approach not only poses significant computational challenges in terms of training a large number of GFlowNets, but also fails to take advantage of the shared structure present between the sub-problems.

% In addition to the computational considerations of training a GFlowNet for a large number of such sub-problems, even in principle this approach does not exploit the shared structure between the sub-problems. 

% Thus, we can employ reward-conditional GFlowNets to model the family of reward functions by using as the conditioning set $\mathcal{C}$ the $d$-simplex $\Delta^d$ spanned by the preferences $\omega$ over $d$ objectives. % for the possible values of the preference weights $\omega$.
\emph{Reward-conditional GFlowNets}~\citep{bengio2021gflownet} are a generalization of GFlowNets that learn a single conditional policy to \emph{simultaneously} model a family of distributions associated with a corresponding family of reward functions. Let $\mathcal{C}$ denote a set of values $c$, with each $c \in \mathcal{C}$ inducing a unique reward function $R(x|c)$. We can define a \emph{family} of weighted DAGs $\{\mathcal{G}_c = \left(\mathcal{S}_c, \mathcal{E} \right), \enspace c \in \mathcal{C}\}$ which describe the construction of $x\in \mathcal{X}$, with conditioning information $c$ available at all states in $\mathcal{S}_c$. Having $c$ as input allows the policy to learn the shared structure across different values of $c$. We denote $P_F(-| s, c)$ and $P_B(- | s', c)$ as the \emph{conditional} forward and backward policies, $Z(c) = \sum_{x \in \mathcal{X}} R(x | c)$ as the \emph{conditional} partition function and $\pi(x|c)$ as the marginal likelihood (given $c$) of sampling trajectories $\tau$ from $P_F$ terminating in $x$. The learning objective in reward-conditional GFlowNets is thus estimating $P_F(-|s, c)$ such that $\pi(x|c)\propto R(x|c)$. Exploiting the shared structure enables a single conditional policy (e.g., a neural net taking $c$ and $s$ as input and actions probabilities as outputs) to model the entire family of reward functions simultaneously. Moreover, the policy can generalize to values of $c$ not seen during training. %We refer the reader to~\citet{bengio2021gflownet} for a more formal discussion of conditional GFlowNets. 

The MOO sub-problems possess a similar shared structure, induced by the preferences. Thus, we can leverage a single reward-conditional GFlowNet instead of a set of independent GFlowNets to model the sub-problems simultaneously. Formally, \emph{Preference-conditional GFlowNets} (MOGFN-PC) are reward-conditional GFlowNets with the preferences $\omega\in\Delta^d$ over the set of objectives $\{R_1(x), \dots,R_d(x)\}$ as the conditioning variable. MOGFN-PC models the family of reward functions defined by the scalarization function $R(x|\omega)$. MOGFN-PC is a general approach and can accommodate any scalarization function, be it existing ones discussed in Section~\ref{sec:moo_bg} or novel scalarization functions designed for a particular task. To illustrate this flexibility, we introduce the Weighted-log-sum (WL): $R(x|\omega) = \prod_{i=1}^d R_i(x)^{\omega_i}$ scalarization function. We hypothesize that the weighted sum in $\log$ space can potentially can help in scenarios where \emph{all} objectives are to be optimized simultaneously, and the scalar reward from Weighted-Sum can be dominated by a single reward. The scalarization function is a key component for MOGFN-PC, and we further study the empirical impact of various scalarization functions in \cref{sec:analysis}.

\paragraph{Training MOGFN-PC}
The procedure to train MOGFN-PC, or any reward-conditional GFlowNet, closely follows that of a standard GFlowNet and is described in \Cref{algo:mogfn-pc}. The objective is to learn the parameters $\theta$ of the forward and backward conditional policies $P_F(- | s, \omega; \theta)$ and $P_B(- | s', \omega;\theta)$, and the log-partition function $\log Z_\theta(\omega)$ such that $\pi(x|\omega)\propto R(x|\omega)$. To this end, we consider an extension of the trajectory balance objective:
\begin{equation}
    \mathcal{L}(\tau, \omega;\theta) = \left(\log \frac{Z_\theta(\omega)\prod_{s\rightarrow s' \in \tau}P_F(s' | s, \omega;\theta)}{R(x | \omega)\prod_{s\rightarrow s' \in \tau}P_B(s | s', \omega;\theta)}\right)^2.
    \label{eq:mogfn_tb}
\end{equation}
One important component is the distribution $p(\omega)$ used to sample preferences during training. $p(\omega)$ influences the regions of the Pareto front that are captured by MOGFN-PC. In our experiments, we use a Dirichlet$(\alpha)$ to sample preferences $\omega$ which are encoded with thermometer encoding \citep{buckman2018thermometer} when input to the policy in some of the tasks. Following prior work, we apply an exponent $\beta$ for the reward $R(x|\omega)$, i.e. $\pi(x|\omega) \propto R(x|\omega)^\beta$. This incentivizes the policy to focus on the modes of $R(x|\omega)$, which is critical for generation of \emph{high reward} diverse candidates. By changing $\beta$ we can trade-off the diversity for higher rewards.  We study the impact of these choices empirically in \cref{sec:analysis}.

\paragraph{MOGFN-PC and MOReinforce}
MOGFN-PC is closely related to MOReinforce~\citep{lin2021pareto} in that both learn a preference-conditional policy to sample Pareto-optimal candidates. The key difference is the learning objective: MOReinforce uses a multi-objective variant of REINFORCE~\citep{williams1992simple}, whereas MOGFN-PC uses a preference-conditional GFlowNet objective (\Cref{eq:mogfn_tb}). MOReinforce, given a preference $\omega$ will converge to generating a single candidate that maximizes $R(x|\omega)$. MOGFN-PC, on the other hand, samples from $R(x|\omega)$, resulting in generation of diverse candidates from the Pareto set according to the preferences specified by $\omega$. This ability to generate diverse Pareto optimal candidates is a key feature of MOGFN-PC, whose advantage is demonstrated empirically in Section~\ref{sec:emp_results}.

\subsection{Multi-Objective Active Learning with GFlowNets}
\label{sec:mogfn-al}
In many applications, the objective functions of interest are often computationally expensive to evaluate. Consider the drug discovery scenario, where evaluating objectives such as the binding energy of a candidate molecule to a target even in imperfect simulations can take several hours. Sample efficiency, in terms of number of evaluations of the objective functions, is therefore critical. 
% Black-box optimization approaches involving active learning~\citep{zuluaga2013active}, particularly multi-objective Bayesian optimization (MOBO) methods~\citep{shah2016pareto,garnett_bayesoptbook_2022} are powerful approaches in these settings. 

We introduce \emph{MOGFN-AL} which leverages GFlowNets to generate candidates in each round of a multi-objective Bayesian optimization loop. MOGFN-AL tackles MOO through a sequence of single-objective sub-problems defined by acquisition function $a$. MOGFN-AL is a multi-objective extension of GFlowNet-AL~\citep{jain2022biological}. Here, we apply MOGFN-AL for biological sequence design summarized in~\Cref{algo:mogfn-al} (\Cref{app:algorithms}), building upon the framework proposed by~\citet{stanton2022lambo}. This problem was previously studied by~\citet{seff2019discrete} and has connections to denoising autoencoders~\citep{bengio2013generalized}.

% We start with an initial dataset $\mathcal{D}_0 = {(x_i, y_i)}_{i=1}^N$ of candidates $x_i\in \mathcal{X}$ and their evaluation with the true objectives $y_i = \mathbf{R}(x)$. $\mathcal{D}_i$ is used to train a surrogate probabilistic model (proxy) of the true objectives $\hat{f}: \mathcal{X} \to \mathbb{R}^d$, which we parameterize as a multi-task Gaussian process~\citep{shah2016pareto} with a deep kernel~\citep[DKL GP;][]{maddox2021bayesian,maddox2021conditioning}. Using this proxy, the acquisition function defines the utility to be maximized $a:\mathcal{X}\times\mathcal{F}\to \mathbb{R}$, where $\mathcal{F}$ denotes the space of functions represented by DKL GPs. In our work we use as acquisition function $a$ \emph{noisy expected hypervolume improvement}~\citep[NEHVI;][]{daulton2020differentiable}.

In existing work applying GFlowNets for biological sequence design, the GFlowNet policy generates the sequences token-by-token~\citep{jain2022biological}. While this offers greater flexibility to explore the space of sequences, it can be prohibitively slow when the sequences are large. %In contrast, 
% In our application of biological sequence design using MOGFN-AL, we adopt an alternative approach inspired by~\citep{stanton2022lambo}. 
In contrast, we use GFlowNets to propose candidates at each round $i$ by generating mutations for existing candidates $x \in \hat{\mathcal{P}}_i$ where $\hat{\mathcal{P}}_i$ is the set of non-dominated candidates in $\mathcal{D}_i$. Given a sequence $x$, the GFlowNet, through a sequence of stochastic steps, generates a set of mutations $m = \{(l_i, v_i)\}_{i=1}^T$ where $l\in \{1, \dots, |x|\}$ is the location to be replaced and $v\in \mathcal{A}$ is the token to replace $x[l]$ while $T$ is the number of mutations. 
% This set is generated sequentially such that each mutation is sampled from $P_F$ conditioned on $x$ and the mutations sampled so far $\{(l_i, v_i)\}$. 
Let ${x'}_m$ be the sequence resulting from mutations $m$ on sequence $x$. The reward for a set of sampled mutations for $x$ is the value of the acquisition function on ${x'}_m$, $R(m, x) = a({x'}_m|\hat{f})$. This mutation based approach scales better to tasks with longer sequences while still affording ample exploration in sequence space for generating diverse candidates. We demonstrate this empirically in~\Cref{sec:al_exp}. %that generating mutations with GFlowNets results in more diverse candidates and faster improvements to the Pareto front than LaMBO~\citep{stanton2022lambo}.

%% file: tex/related.tex
% \vspace{-4mm}
\section{Related Work}
\label{sec:related}
% \vspace{-4mm}
%Multi-objective optimization has been a long-standing challenge for many real-world challenges with a variety of algorithms in adjacent fields having been developed in recent years. 

% \textbf{Evolutionary Algorithms:} Many long-standing approaches showcased in \citet{Konak2006} continue to serve as useful algorithms applied to this day in a variety of engineering and research challenges. Recently, many classic evolutionary algorithms have been integrated into a user-friendly library in \citet{pymoo}, allowing researchers to quickly test algorithms for their desired problems. Furthermore, recent approaches like \citet{miret2022neuroevolution} have integrated specialized neural network architectures, such as graph convolutional networks, into the framework of traditional evolutionary algorithms and thereby increased their ability to resolve very large combinatorial search spaces. Our \ac{MOGFN} formulation fundamentally differs from the evolutionary algorithm approaches by leveraging prior experience to construct a policy users can sample from leading to broader generalizability. 
\paragraph{Evolutionary Algorithms (EA)} Traditionally, evolutionary algorithms such as NSGA-II have been widely used in various multi-objective optimization problems~\citep{ehrgott2005multicriteria,Konak2006,pymoo}. More recently, \citet{miret2022neuroevolution} incorporated graph neural networks into evolutionary algorithms enabling them to tackle large combinatorial spaces. Unlike MOGFNs, evolutionary algorithms are required to solve each instance of a MOO from scratch rather than by amortizing computation during training in order to quickly generate solutions at run-time. Evolutionary algorithms, however, can be augmented with MOGFNs for generating mutations to improve efficiency, as in \Cref{sec:mogfn-al}.  %\textcolor{red}{TODO: How these methods contrast MOGFN?}
\paragraph{Multi-Objective Reinforcement Learning}MOO problems have also received significant interest in the RL literature~\citep{hayes2022practical}. Traditional approaches broadly consist of learning sets of Pareto-dominant policies~\citep{roijers2013survey,van2014multi,reymond2022pareto}. Recent work has focused on extending Deep RL algorithms for multi-objective settings, e.g., with Envelope-MOQ~\citep{yang2019generalized}, MO-MPO~\citep{abdolmaleki2020distributional,abdolmaleki2021multi} , and MOReinforce~\citep{lin2021pareto}. A general shortcoming of RL-based approaches is their objective focuses on discovering a single mode of the reward function, and thus hardly generate diverse candidates, an issue that also persists in the multi-objective setting.  In contrast, MOGFNs sample candidates proportional to the reward, implicitly resulting in diverse candidates.
% \textbf{Multi-Objective Bayesian Optimization:} In order to define expected improvement metrics important for Bayesian optimization algorithms, many recent works leverage multi-objective utility functions, such as expected hypervolume improvement \citep{emmerich2011hypervolume} and its subsequent improvements \citep{daulton2020differentiable, daulton2021parallel} that have multi-objective Bayesian optimization to scale a significant number of objectives for challenging engineering problems \citep{daulton2022multiobjective}. As an alternative to multi-objective utility functions, prior works have also conditioned expected improvement on preferences for different objectives \citep{abdolshah2019multi}, or applied scalarization techniques \citep{paria2020flexible, zhang2020random} to reformulate the multi-objective problem for Bayesian optimization. Our GFlowNet formulation leverages ideas similar to those in the aforementioned works, as it has the flexibility to condition on different objective preferences and optimize multiobjective utility metrics. As mentioned in the multi-objective RL description, \ac{MOGFN}s have the additional advantage of naturally providing diverse solutions in the input space of the problem, which is also an advantage compared to multi-objective Bayesian optimization.
\paragraph{Multi-Objective Bayesian Optimization (MOBO)} Bayesian optimization (BO) has been used in the context of MOO when the objectives are expensive to evaluate and sample efficiency is a key consideration. MOBO approaches consist of learning a surrogate model of the true objective functions, which is used to define an acquisition function such as expected hypervolume improvement~\citep{emmerich2011hypervolume,daulton2020differentiable,daulton2021parallel} and max-value entropy search~\citep{belakaria2019max}, as well as scalarization-based approaches~\citep{paria2020flexible,zhang2020random}. \citet{abdolshah2019multi} and \citet{lin2022preference} study the MOBO problem in the setting with preferences over the different objectives. \citet{stanton2022lambo} proposed LaMBO, which uses language models in conjunction with BO for multi-objective sequence design problems. While recent work~\citep{konakovic2020diversity,maus2022discovering} studies the problem of generating diverse candidates in the context of MOBO, it is limited to local optimization near Pareto-optimal candidates in low-dimensional continuous problems. As such, the key drawbacks of MOBO approaches are that they typically do not consider the need for diversity in generated candidates and that they mainly consider continuous low-dimensional state spaces. As we discuss in~\Cref{sec:mogfn-al}, MOBO approaches can be augmented with GFlowNets for \textit{diverse} candidate generation in discrete spaces. 

\paragraph{Other Approaches} \citet{zhao2022multiobjective} introduced LaMOO which tackles the MOO problem by iteratively splitting the candidate space into smaller regions, whereas \citet{daulton2022multiobjective} introduce MORBO, which performs BO in parallel on multiple local regions of the candidate space. Both these methods, however, are limited to continuous candidate spaces. 
% \subsection{Evolutionary Algorithms}
% \citep{Konak2006, miret2022neuroevolution,pymoo}

% \subsection{Reinforcement Learning}
% DQN-Extension (Envelope Q-Learning):
% \citep{yang2019generalized}
% MO-Reinforce:
% \citep{lin2021pareto}
% Review Paper
% \citep{hayes2022practical}
% Novel Approaches with conditioning and Pareto Partitioning
% \citep{reymond2022pareto,navon2021learning}\citep{zhao2022multiobjective}

% Pareto Set Learning and Single Objective Learning
% \citep{ren2020solving}

% \subsection{Bayesian Optimization}
% \citep{balandat2020botorch,daulton2020differentiable,daulton2021parallel,stanton2022lambo}
% \citep{jin2020multi,daulton2022multiobjective}
% \citep{abdolshah2019multi}
% \cite{zhang2020random,paria2020flexible}

% \subsection{Multi-Task Learning}
% \cite{sener2018multi}

%% file: tex/empirical_results.tex
% \vspace{-4mm}
\section{Empirical Results}
\label{sec:emp_results}
\begin{figure*}[ht!]
     \centering
     \begin{subfigure}[b]{0.4\textwidth}
         \centering
         \includegraphics[width=\textwidth]{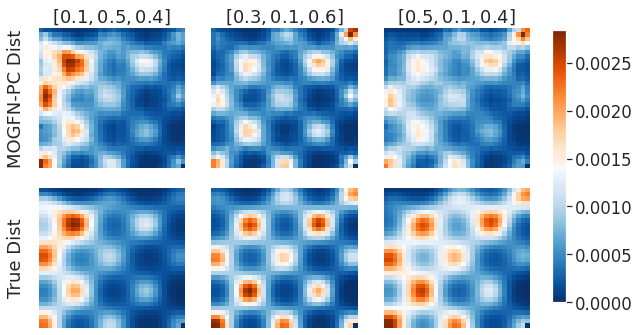}
         \caption{}
         \label{fig:grid_qualitative}
     \end{subfigure}
    %  \hfill
     \begin{subfigure}[b]{0.25\textwidth}
         \centering
         \includegraphics[width=0.85\textwidth]{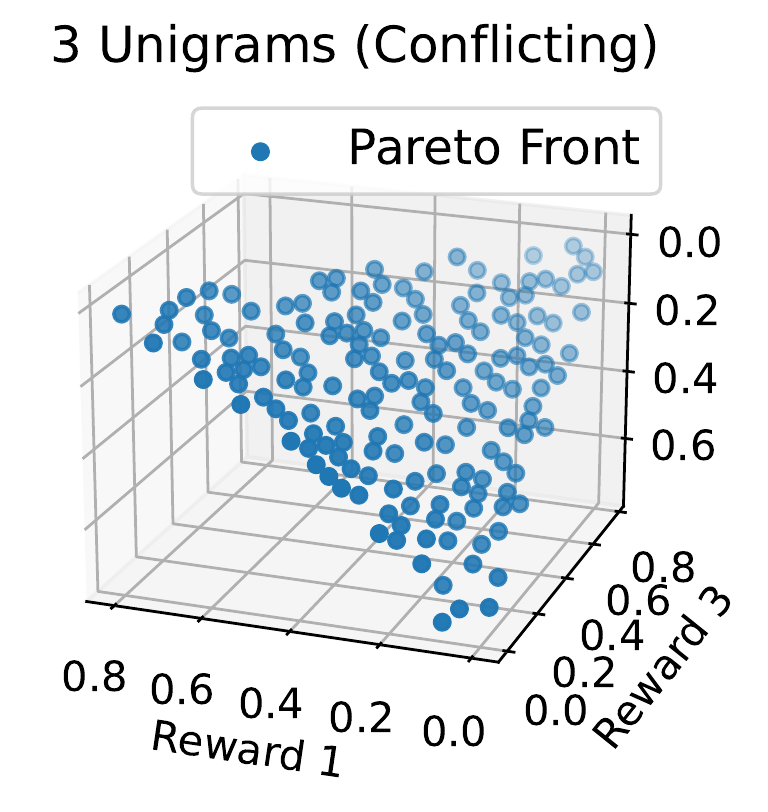}
         \caption{}
         \label{fig:string_conf_front}
     \end{subfigure}
     \begin{subfigure}[b]{0.25\textwidth}
         \centering
         \includegraphics[width=0.85\textwidth]{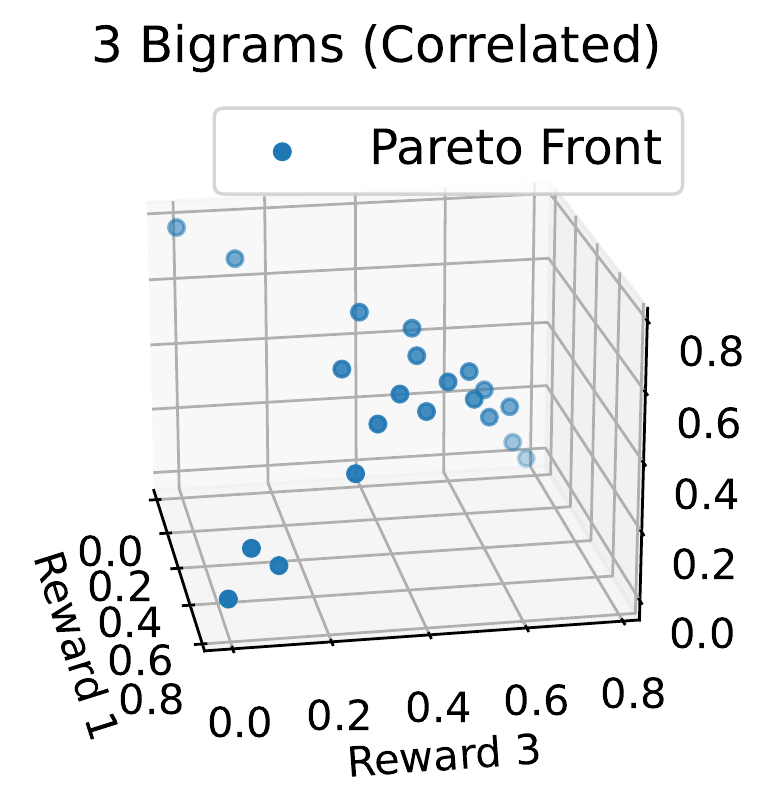}
         \caption{}
         \label{fig:string_cor_front}
     \end{subfigure}
        \caption{\textit{(a)} The distribution learned by MOGFN-PC (Top) almost exactly matches the ground truth distribution (Bottom), in particular capturing all the modes, on hypergrid of size $32 \times 32$ with 3 objectives. 
        % Top row corresponds to the MOGFN distribution for a given preference vector $\omega$ after training and the bottom row is the true distribution. Note that the MOGFN captures all the modes in the distribution. 
        \textit{(b)} and \textit{(c)} Pareto front of candidates generated by MOGFN-PC on the N-grams task illustrates that the MOGFN-PC can model conflicting and correlated objectives well.} 
        \label{fig:grid_results_alt}
        \vspace{-6mm}
\end{figure*}

% \vspace{-4mm}
% \begin{figure}[t]
%     \centering
%     \includegraphics[width=\columnwidth]{fig/MO-GFN-tasks-8-1-1.png}
%     \caption{\textbf{Tasks considered}: We consider two classes of tasks where the GFlowNet graph $\mathcal{G}$ is a general DAG (left) or a tree (right). One-step transition from the current \textit{State} to  a \textit{New State}. \textit{Left}: All the molecular tasks considered in this paper (QM9 and Fragments) presume a DAG structure. \textit{Right}: All the sequence modelling tasks (Strings and DNA generation) presume a tree structure.}
%     \label{fig:tasks}
% \end{figure}

In this section, we present our empirical findings across a wide range of tasks ranging from sequence design to molecule generation. 
%The experiments cover two distinct classes of problems in the context of GFlowNets: where $\mathcal{G}$ is a DAG and where it is a tree.
% We consider two classes of tasks as summarized in \Cref{fig:tasks}: Graph generation where the trajectories $\tau$ induce a DAG (left) and sequence generation where the trajectories $\tau$ induce a tree (right).
Through our experiments, we aim to answer the following questions:
\vspace{-3mm}
\begin{enumerate}[leftmargin=6.5mm]
\setlength\itemsep{0.1pt}
    \item [\textbf{Q1}] \emph{Can MOGFNs model the preference-conditional reward distribution?}
    \item [\textbf{Q2}] \emph{Can MOGFNs sample Pareto-optimal candidates?}
    \item [\textbf{Q3}] \emph{Are candidates sampled by MOGFNs \emph{diverse}?}
    \item [\textbf{Q4}] \emph{Do MOGFNs scale to high-dimensional problems relevant in practice?}
\end{enumerate}
\vspace{-3mm}
We obtain positive experimental evidence for \textbf{Q1-Q4}.

\textbf {Metrics:} %To answer these questions we rely on standard metrics such as \textbf{Hypervolume} (HV), \textbf{Generational Distance+} (GD+), and $\boldsymbol{R}_2$ \textbf{indicator} to quantify the quality of the approximate Pareto front. 
We rely on standard MOO metrics such as the \textbf{Hypervolume} (HV) and $\boldsymbol{R}_2$ \textbf{indicators}, as well as the \textbf{Generational Distance+} (GD+). 
To measure diversity we use the \textbf{Top-K Diversity} and \textbf{Top-K Reward} metrics of~\citet{bengio2021flow}. We detail all metrics in Appendix~\ref{app:metrics}. For all our empirical evaluations we follow the same protocol.  First, we sample a set of preferences which are fixed for all the methods. For each preference we sample $128$ candidates from which we pick the top $10$, compute their scalarized reward and diversity, and report the averages over preferences.
We then use these samples to compute the HV and $R_2$ indicators. %, we pick the best candidate based on the weighted-sum aggregation per preference and average the scores across all preferences. Also, 
We pick the best hyperparameters for all methods based on the HV and report the mean and standard deviation over $3$ seeds for all quantities.

\textbf{Baselines}: We consider the closely related MOReinforce~\citep{lin2021pareto} as a baseline. We also study its variants MOSoftQL and MOA2C which use Soft Q-Learning~\citep{haarnoja2017reinforcement} and A2C~\citep{mnih2016asynchronous} in place of REINFORCE.
% as it also learns a preference-conditional policy using REINFORCE algorithm to sample Pareto-optimal candidates. 
We additionally compare against Envelope-MOQ~\citep{yang2019generalized}, another popular multi-objective reinforcement learning method. 
For fragment-based molecule generation we consider an additional baseline, MARS~\citep{xie2021mars}, a relevant MCMC approach for this task. Notably, we do not consider baselines like LaMOO~\citep{zhao2022multiobjective} and MORBO~\citep{daulton2022multiobjective} as they are designed for continuous spaces and rely on latent representations from pre-trained models for discrete tasks like molecule generation, making the comparisons unfair as the rest of the methods are trained on significantly fewer molecules. Additionally, it is not clear to apply them on tasks like DNA Aptamer design, where pretrained models are not available. 
% For MOGFN-AL experiments we consider LaMBO baseline \cite{stanton2022lambo} which uses denoising autoencoder and a gaussian process head for multi-objective optimization in the latent space. 

%% file: tex/grid.tex
\subsubsection{Hyper-Grid}

\label{sec:grid}
We first study the ability of MOGFN-PC to capture the preference-conditional reward distribution in a multi-objective version of the HyperGrid task from \citet{bengio2021flow}. The goal here is to navigate proportional to a reward within a HyperGrid. We consider the following objectives for our experiments: $\texttt{brannin(x), currin(x), shubert(x)}$\footnote{We present additional results with more objectives in \Cref{sec:app-hypergrid}}.

Since the state space is small, we can compute the distribution learned by MOGFN-PC in closed form. In \Cref{fig:grid_qualitative}, we visualize $\pi(x|\omega)$, the distribution learned by MOGFN-PC conditioned on a set of fixed preference vectors $\omega$ and contrast it with the true distribution $R(x|\omega)$ in a $32\times 32$ hypergrid with 3 objectives. We observe that $\pi(-|\omega)$ and $R(-|\omega)$ are very similar. %the learned distribution is almost the same as the true distribution. 
To quantify this, we compute $\mathbb{E}_x\left[|\pi(x|\omega)-R(x|\omega)/Z(\omega)|\right]$ averaged over a set of $64$ preferences, and find a difference of about $10^{-4}$. Note that MOGFN-PC is able to capture all the modes in the distribution, which suggests the candidates sampled from $\pi$ would be diverse. Further, we compute the GD+ metric for the Pareto front of candidates generated with MOGFN-PC, which comes up to an average value of 0.42. For more details % and additional experiments studying varying number of objectives please 
about the task and the additional results, refer to \Cref{sec:app-hypergrid}.

% Apart from this, the average $L_1$-loss between the MOGFN-PC distribution and the true distribution averaged across a set of fixed test preferences is  . We can see that for all objectives considered the loss is consistently low. Finally, to test the Pareto performance, we measure the \textbf{Generational Distance $+$} (GD$+$) across different numbers of objectives. We observe that the MOGFN-PC not only learns a distribution that is close to the true distribution but also learns to sample from the Pareto front. 

% alt version - attempt at compressing
% \Cref{fig:grid_results} illustrates the conditional distribution learnt by MOGFN-PC $\pi(x|\omega)$ for a set of preferences $\omega$, along with the true conditional reward distribution $R(x|\omega)$ on a $32 \times 32$ hypergrid with 3 objectives. We answer \textbf{Q1} affirmatively, as the distribution learned by \ac{MOGFN}-PC $\pi(x|\omega)$ matches $R(x|\omega)$ almost exactly. Moreover, \ac{MOGFN}-PC is able to capture all the modes in the distribution, which suggests that diversity is captured, answering \textbf{Q3}. 

%% file: tex/string.tex
\subsubsection{N-grams Task}
\label{sec:string}
We consider version of the synthetic sequence design task from \citet{stanton2022lambo}. The task consists of generating strings with the objectives given by occurrences of a set of $d$ n-grams. 
% See \Cref{sec:app-string} for more details on the task.
% Given an alphabet of available actions, the task is to generate sequences (of maximum length $32$) which maximize the set of objectives $\{R_1, \dots, R_d\}$. Each reward $R_i$ is defined as the number of occurrences of a given pattern in the sequence. 
% For instance, given the pattern set \texttt{["A", "B", "C"]}, the reward for sequence, $x = "AABCCD"$ would be $[2, 1, 2]$. 
% We consider two tasks: \texttt{regex\_3} with patterns \texttt{["AC", "CV", "VA"]} (correlated objectives, i.e. multiple objectives can be maximized simultaneously) and \texttt{regex\_3\_conf} with patterns \texttt{["A", "C", "V"]} (conflicting rewards, i.e. one objective cannot be improved without making another worse). 
% Note that as the reward depends only on the number of occurrences of the patterns, there can be multiple sequences corresponding to the same reward, making the notion of diversity important. We use the \emph{edit distance} as the measure of distance between two candidates. We present further details in \Cref{sec:app-string}. 
% We use MOReinforce~\citep{lin2021pareto}, Envelope-MOQ~\cite{yang2019generalized} and MOSoftQL (same formulation as MOReinforce and MOGFN-PC, but trained with the Soft Q-Learning~\cite{haarnoja2017reinforcement} objective) as baselines.
MOGFN-PC adequately models the trade-off between conflicting objectives in the 3 Unigrams task as illustrated by the Pareto front of generated candidates in~\Cref{fig:string_conf_front}. For the 3 Bigrams task with correlated objectives (as there are common characters in the bigrams), \Cref{fig:string_cor_front} demonstrates MOGFN-PC generates candidates which can simultaneously maximize multiple objectives. We refer the reader to \Cref{sec:app-string} for more task details and additional results with different number of objectives and varying sequence length.
In the results summarized in \Cref{tab:stringregex3} in the Appendix, MOGFN-PC outperforms the baselines in terms of the MOO metrics while generating diverse candidates. Note that as occurrences of n-grams is the reward, the diversity is limited by the performance, i.e. high scoring sequences will have lower diversity, explaining higher diversity of MOSoftQL. We also observe that the MOReinforce and Envelope-MOQ baselines struggle in this task potentially due to longer trajectories with sparse rewards.

%% file: tex/mol_exp.tex
\subsubsection{QM9}
\label{sec:qm9}
We first consider a small-molecule generation task based on the QM9 dataset~\citep{ramakrishnan2014quantum}. We generate molecules atom-by-atom and bond-by-bond with up to 9 atoms and use 4 reward signals. The main reward is obtained via a MXMNet~\citep{zhang2020molecular} proxy trained on QM9 to predict the HOMO-LUMO gap. The other rewards are Synthetic Accessibility (SA), a molecular weight target, and a molecular logP target. 
Rewards are normalized to be between 0 and 1, but the gap proxy can exceed 1, and so is clipped at 2. The preferences $\omega$ are input to the policy as is, without thermometer encoding as we find no significant difference between the two choices.   %.\!\footnote{Since the gap proxy is an approximate model, it can output higher values than those found in its training set, meaning the gap reward can exceed 1; it is clipped at 2.}
%All rewards are normalized so that they are between 0 and 1.\!\footnote{Since the gap proxy is an approximate model, it can output higher values than those found in its training set, meaning the gap reward can exceed 1; it is clipped at 2.}
% \textbf{Baselines}: We compare GFlowNet against a similarly rewarded preference-conditioned Advantage Actor Critic~\citep{mnih2016asynchronous} agent (which we have found was the best performing \emph{vanilla} RL method), and against an Envelope Q-Learning agent~\citep{yang2019generalized}.
% \begin{itemize}
%     \item \textcolor{red}{PPO Agent - TODO} 
%     \item \textcolor{red}{Flow matching Agent - TODO}
% \end{itemize}
%We present the results in \Cref{tab:qm9_results}. 
We train the models with 1M molecules and present the results in \Cref{tab:qm9_results}, showing that MOGFN-PC outperforms all baselines in terms of Pareto performance and diverse candidate generation. 
% We compare MOGFN against MOReinforce \cite{lin2021pareto}, MOA2C, which is a preference-conditioned based A2C algorithm \cite{mnih2016asynchronous}, and Envelope Q-Learning \cite{yang2019generalized}. 

\begin{table}[ht]
    \caption{\textbf{Atom-based QM9 task:} MOGFN-PC exceeds Diversity and Pareto performance on QM9 task with HUMO-LUMO gap, SA, QED and molecular weight objectives compared to baselines.}
    \centering

\resizebox{0.48\textwidth}{!}{
    \begin{tabular}{ccccc}
    \toprule
     Algorithm   & Reward ($\uparrow$)  & Diversity ($\uparrow$) & HV ($\uparrow$) & $R_2$ ($\downarrow$) \\
    \midrule
Envelope QL & \g{0.65}{0.06} & \g{0.85}{0.01} &\g{1.26}{0.05} & \g{5.80}{0.20}\\
MOReinforce  & \g{0.57}{0.12} & \g{0.53}{0.08} & \g{1.35}{0.01} & \g{4.65}{0.03}\\
MOA2C & \g{0.61}{0.05} & \g{0.39}{0.28} & \g{1.16}{0.08} & \g{6.28}{0.67} \\
MOGFN-PC & \highlight{\g{0.76}{0.00}} & \highlight{\g{0.93}{0.00}} & \highlight{\g{1.40}{0.18}} & \highlight{\g{2.44}{1.88}}\\
\bottomrule
    \end{tabular}
    }
    \label{tab:qm9_results}
    % \vspace{-4mm}
\end{table}

\begin{table}[ht]
    \caption{\textbf{Fragment-based Molecule Generation Task: } Diversity and Pareto performance on the Fragment-based drug design task with sEH, QED, SA and molecular weight objectives.}% The arrows indicate the desirable trend of the metrics.}
    \centering

\resizebox{0.48\textwidth}{!}{
    \begin{tabular}{c c c c c}
    \toprule
     Algorithm   & Reward ($\uparrow$)  & Diversity ($\uparrow$) & HV ($\uparrow$) & $R_2$ ($\downarrow$) \\
    \midrule
    Envelope QL& \g{0.70}{0.10} & \g{0.15}{0.05} &  \g{0.74}{0.01} & \g{3.51}{0.10} \\
    MARS  & -- & -- &  \g{0.85}{0.008} & \g{1.94}{0.03}  \\
    MOReinforce  & \g{0.41}{0.07} & \g{0.01}{0.007} &\g{0}{0} &  \g{9.88}{1.06} \\
    MOA2C & \g{0.76}{0.16} & \g{0.48}{0.39} &  \g{0.75}{0.01} & \g{3.35}{0.02} \\
    MOGFN-PC & \highlight{\g{0.89}{0.05}} & \highlight{\g{0.75}{0.01}} &  \highlight{\g{0.90}{0.01}} & \highlight{\g{1.86}{0.08}}  \\
    \bottomrule
    \end{tabular}
    }
    \label{tab:fragments}
    % \vspace{-6mm}
\end{table}

\subsubsection{Fragment-Based Molecule Generation}
\label{sec:frags}
We evaluate our method on the fragment-based~\citep{kumar2012fragment} molecular generation task of \citet{bengio2021flow}, where the task is to generate molecules by linking fragments to form a junction tree~\citep{Jin_2020}. %This is a graph generation task where the nodes in the graph correspond to heavy atoms and the edges correspond to  the chemical bonds. This task setup was previously considered in \cite{bengio2021flow} for single objective molecular discovery and in this work, we extend their code to multi-objective setting. More specifically, we consider the following rewards. 
The main reward function is obtained via a pretrained proxy, available from~\citet{bengio2021flow}, trained on molecules docked with AutodockVina~\citep{trott2010autodock} for the sEH target. The other rewards are based on Synthetic Accessibility (SA), drug likeness (QED), and a molecular weight target. We detail the reward construction in \Cref{sec:app-fragments}. The preferences $\omega$ are input to the policy directly for this task as well. %We perform experiments with these four objectives and report HV, $R_2$, Top-k reward, and Top-k diversity. 
Similarly to QM9, we train MOGFN-PC to generate 1M molecules and
%and then calculate HV and $R_2$ indicators, as well as Top-K Rewards, and Top-K Diversity. We 
report the results in \Cref{tab:fragments}. We observe that MOGFN-PC is consistently outperforming baselines not only in terms of HV and $R_2$, but also candidate diversity score. Note that we do not report reward and diversity scores for MARS, since the lack of preference conditioning would make it an unfair comparison.

%% file: tex/dna.tex
\subsubsection{DNA Sequence Generation}
\label{sec:dna-exps}
A practical example of a design problem where the GFlowNet graph is a tree is the generation of DNA aptamers, single-stranded nucleotide sequences that are popular in biological polymer design due to their specificity and affinity as sensors in crowded biochemical environments \citep{zhou2017dnametal,corey2022dnatherapeutics,yesselman2019rnadesign,kilgour2021e2edna}. We generate sequences by adding one nucleobase (A, C, T or G) at a time, with a maximum length of 60 bases. We consider three objectives: the free energy of the secondary structure calculated with the software NUPACK \citep{zadeh2011nupack}, the number of base pairs and the inverse of the sequence length (to favour shorter sequences). 
% We use MOReinforce~\citep{lin2021pareto} as the baseline for comparison.

% As in the other tasks, we evaluate \ac{MOGFN}-PC and the baselines by calculating the hypervolume, Top-K rewards, Top-K diversity and $R_2$ distance. 
The results summarized in Table~\ref{tab:dna_res} indicate that MOReinforce achieves the best MOO performance. However, we note that it finds a quasi-trivial solution with the pattern \texttt{GCGCGC...} for most lengths, yielding low diversity. In contrast, MOGFN-PC obtains much better diversity and Top-K rewards with slightly lower Pareto performance. See \Cref{sec:app-dna} for further discussion.

\begin{table}[h]
    \caption{ \textbf{DNA Sequence Design Task: }Diversity and Pareto performance of various algorithms on DNA sequence generation task with free energy, number of base pairs and inverse sequence length objectives. %The arrows indicate the desirable trend of the metrics.
    % Metrics obtained in the DNA sequence generation task by \ac{MOGFN}-PC with different reward exponent $\beta$ and the reinforcement learning baseline.}
    }
    \centering

\resizebox{0.48\textwidth}{!}{
    \begin{tabular}{ccccc}
    \toprule
   Algorithm & Reward ($\uparrow$) & Diversity ($\uparrow$) & HV ($\uparrow$) & $R_2$ ($\downarrow$) \\
    \midrule
    Envelope-MOQ & \g{0.238}{0.042} & \g{0.0}{0.0} & \g{0.163}{0.013} & \g{5.657}{0.673} \\
    MOReinforce & \g{0.105}{0.002} & \g{0.6178}{0.209} & \highlight{\g{0.629}{0.002}} & \highlight{\g{1.925}{0.003}} \\
    MOSoftQL & \g{0.446}{0.010} & \highlight{\g{32.130}{0.542}} & \g{0.163}{0.014} & \g{5.565}{0.170} \\
    MOGFN-PC & \highlight{\g{0.682}{0.021}} & \g{18.131}{0.981} & \g{0.517}{0.006} & \g{2.432}{0.002} \\
    \bottomrule
    \end{tabular}
    }
    \label{tab:dna_res}
\end{table}

%% file: tex/al.tex
\subsection{Active Learning}
\label{sec:al_exp}

\begin{figure}[ht]% {r}{0.5\textwidth}
% \vspace{-6mm}
  	\centering
    \begin{subfigure}[b]{0.235\textwidth}
        \includegraphics[width=1.0\textwidth]{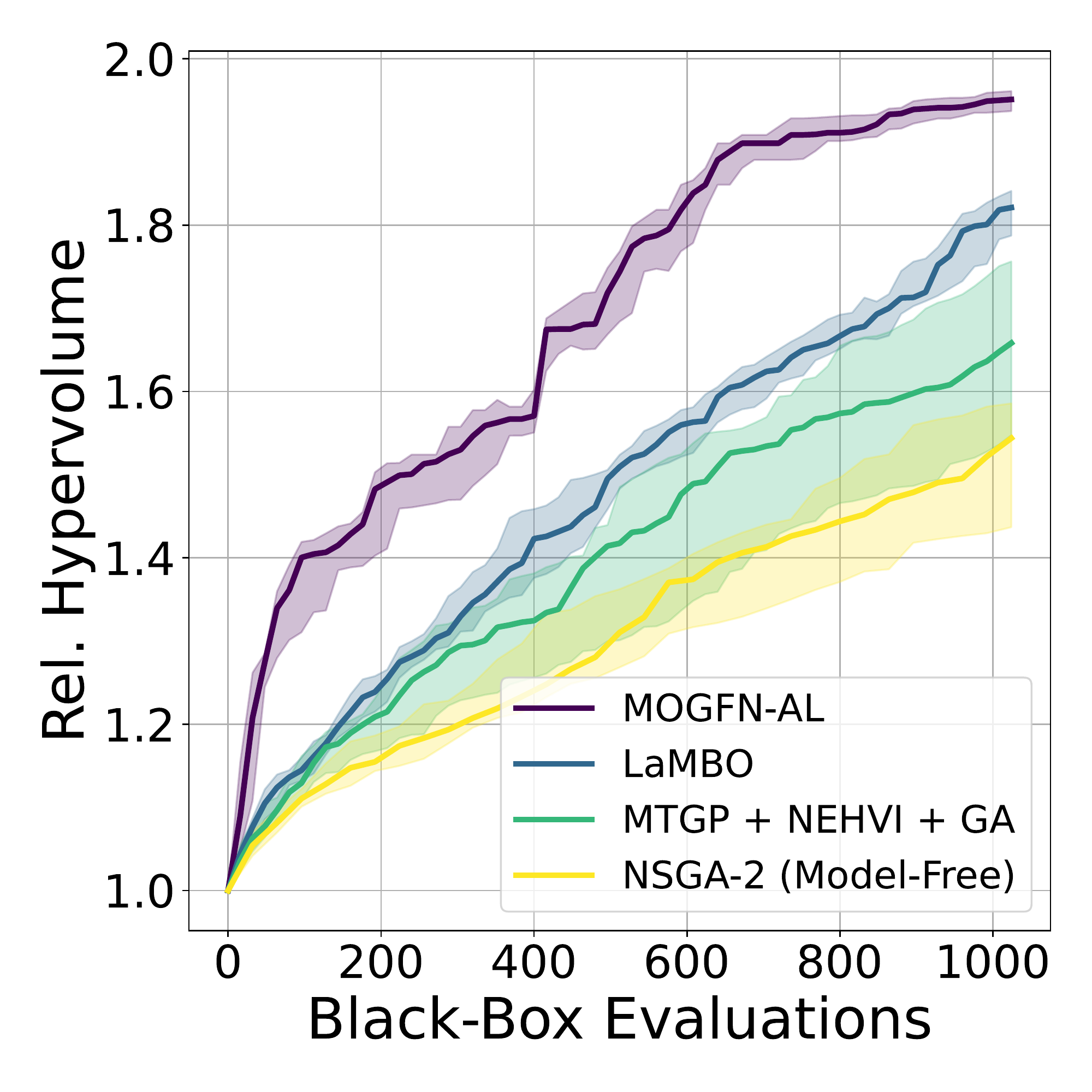}
        \caption{}
        \label{fig:al_hv}
    \end{subfigure}
    \begin{subfigure}[b]{0.235\textwidth}
        \includegraphics[width=1\textwidth]{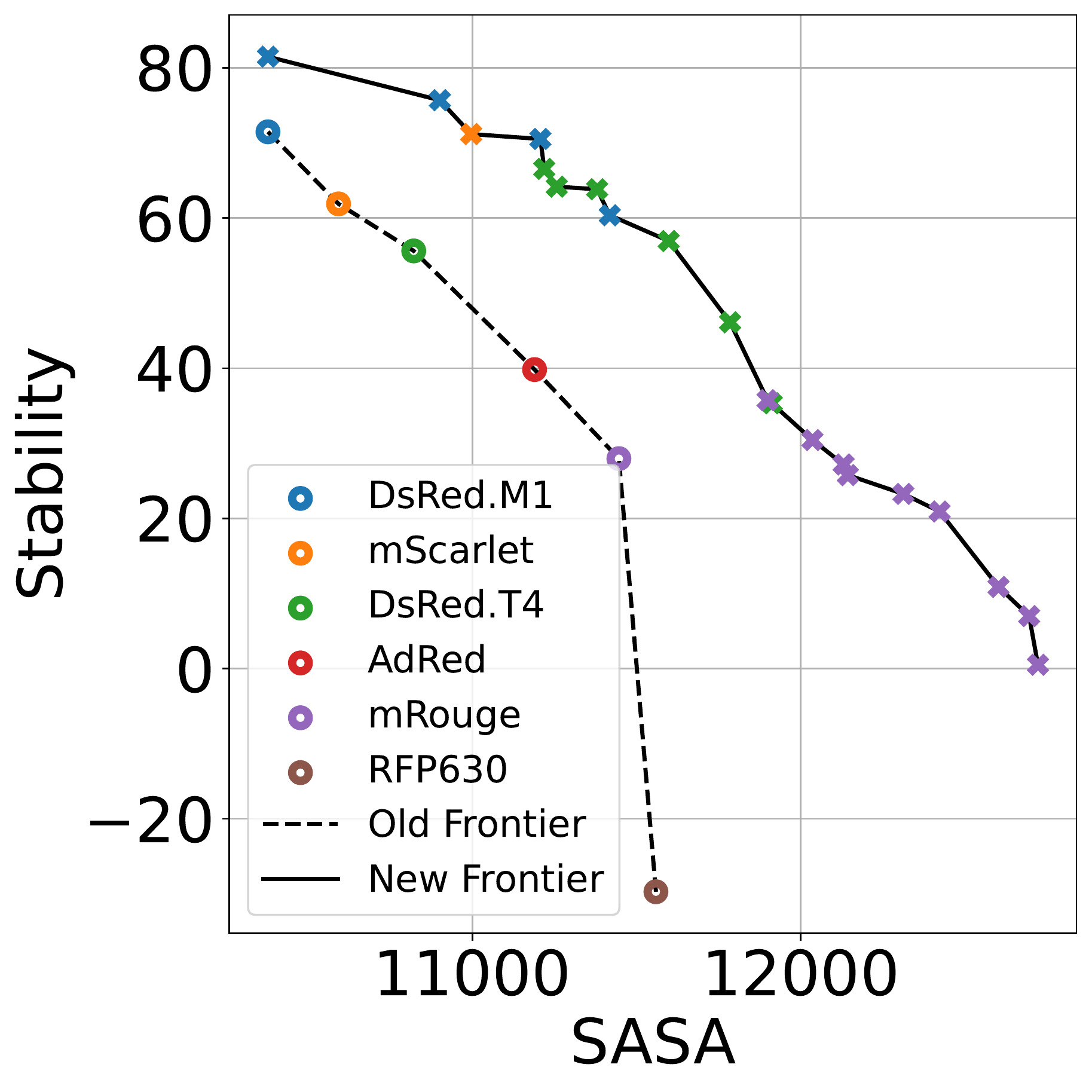}
        \caption{}
        \label{fig:al_pareto}
    \end{subfigure}

\begin{subfigure}[b]{0.3\textwidth}
    \resizebox{1\textwidth}{!}{
    \begin{tabular}{c c}
    \toprule
  Algorithm  & Diversity ($\uparrow$) \\
    \midrule
    NSGA-2 & \g{0.07}{0.03}  \\
    MTGP + NEHVI + GA & \g{0.12}{0.02} \\
    LaMBO & \g{0.18}{0.03} \\
    MOGFN-AL & \highlight{\g{0.25}{0.01}} \\
    \bottomrule
    \end{tabular}
    }
\caption{}
\label{tab:al_diversity}
\end{subfigure}		
    
    \caption{MOGFN-AL demonstrates a substantial advantage in terms of (a) Relative Hypervolume, and (b) the Pareto frontier of candidates generated by MOGFN-AL dominates the Pareto front of the initial dataset, while being more diverse (c) than the baselines'.}
\vspace{-4mm}
\end{figure}

Finally, to evaluate MOGFN-AL, we consider the Proxy RFP task from~\citet{stanton2022lambo}, with the aim of discovering novel proteins with red fluorescence properties, optimizing for folding stability and solvent-accessible surface area. We adopt all the experimental details (described in~\Cref{sec:app-activelearning}) from~\citet{stanton2022lambo}, using MOGFN-AL for candidate generation. In addition to LaMBO, we use a model-free (NSGA-2) and model-based EA from~\citet{stanton2022lambo} as baselines. 
% We use the improvement in HV relative to the initial dataset and the diversity of generated candidates, quantified by the average pairwise BLAST score (e-value). 
We observe in \Cref{fig:al_hv} that MOGFN-AL results in significant gains to the improvement in Hypervolume relative to the initial dataset, in a given budget of black-box evaluations. In fact, MOGFN-AL is able to match the performance of LaMBO within about half the number of black-box evaluations.

\Cref{fig:al_pareto} illustrates the Pareto frontier of candidates generated with MOGFN-AL, which dominates the Pareto frontier of the initial dataset. As we the candidates are generated by mutating sequences in the existing Pareto front, we also highlight the sequences that are mutations of each seqeunce in the initial dataset with the same color. To quantify the diversity of the generated candidates we measure the average e-value from DIAMOND~\citep{buchfink2021sensitive} between the initial Pareto front and the Pareto frontier of generated candidates. Table~\ref{tab:al_diversity} shows that MOGFN-AL generates candidates that are more diverse than the baselines. 

%% file: tex/analysis.tex
\section{Analysis}
\label{sec:analysis}
\begin{figure}[h]% {r}{0.5\textwidth}
% \vspace{-6mm}
%   	\centering
    \begin{subfigure}[t]{0.48\textwidth}
        \centering
        \includegraphics[width=0.8\textwidth]{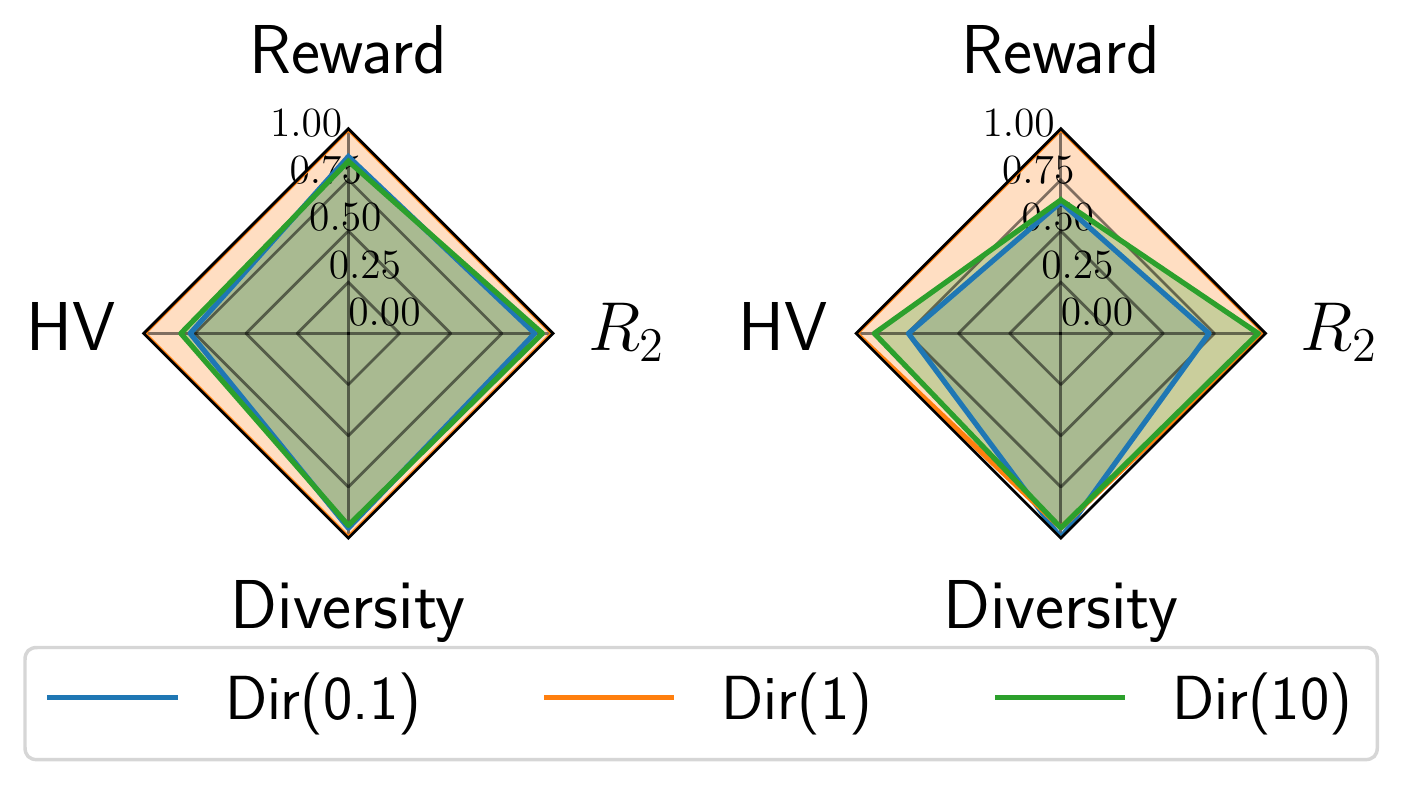}
        % \hfill
        % \includegraphics[width=0.45\textwidth]{fig/abl_mol_prior.pdf}
        \caption{$p(\omega)$}
        \label{fig:ablation_prior}
    \end{subfigure}
    \vfill
    \begin{subfigure}[t]{0.48\textwidth}
        \centering
        \includegraphics[width=0.8\textwidth]{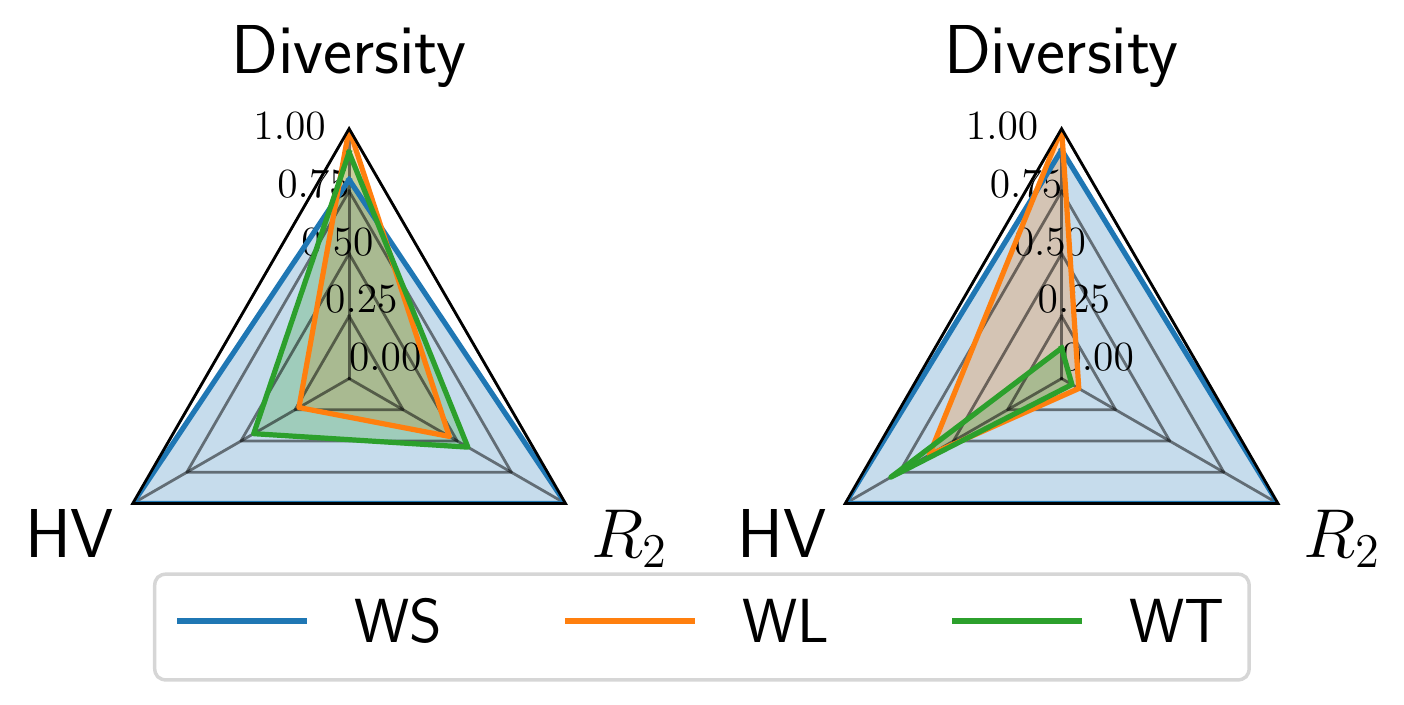}
        % \vfill
        % \includegraphics[width=1\textwidth]{fig/abl_mol_scalarization.pdf}
        \caption{$R(x|\omega)$}
        \label{fig:ablation_scalar_fn}
    \end{subfigure}
    \vfill
    \begin{subfigure}[t]{0.48\textwidth}
        \centering
        \includegraphics[width=0.8\textwidth]{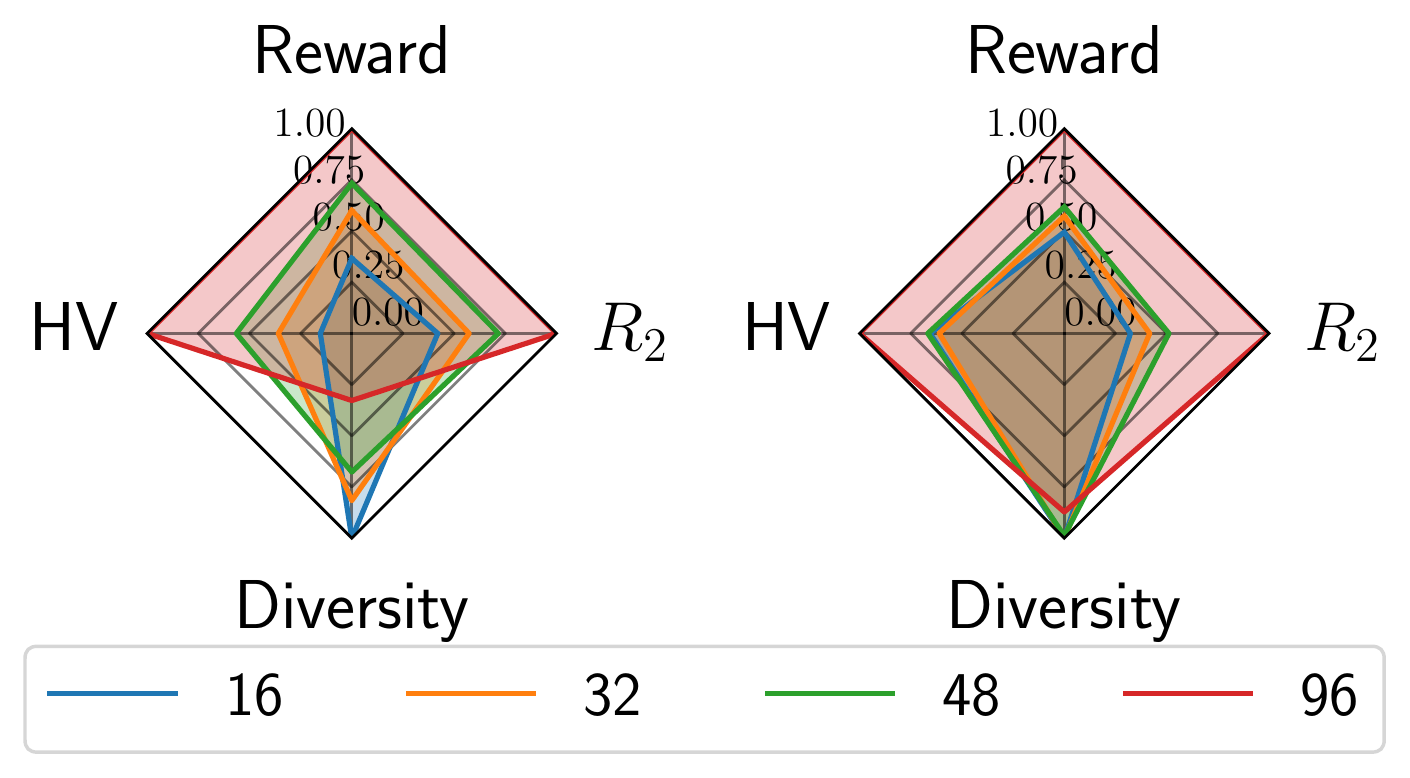}
        % \vfill
        % \includegraphics[width=1\textwidth]{fig/abl_mol_beta.pdf}
        \caption{$\beta$}
        \label{fig:ablation_beta}
    \end{subfigure}
    \caption{Analysing the effect of $p(\omega)$, $R(x|\omega)$ and $\beta$ in the 3 Bigrams (left) and fragment-based molecule generation (right) tasks. The metrics are normalized by the maximum value obtained. As key takeaways, Dirichlet($\alpha=1$) for $p(\omega)$ and weighted sum (WS) scalarization have the best performance in both tasks while the choice of $\beta$ is task-dependent.}
    \label{fig:ablation}
\end{figure}

In this section, we isolate the important components of MOGFN-PC: the distribution $p(\omega)$ for sampling preferences during training, the reward exponent $\beta$ and the scalarization function $R(x|\omega)$ to understand their impact on the performance. \Cref{fig:ablation} summarizes the results of the analysis on the 3 Bigrams task (\Cref{sec:string}) and the fragment-based molecule generation task (\Cref{sec:qm9}) with additional results in the \cref{app:additional_ablations}.

\paragraph{Impact of $p(\omega)$.} 
% The preferences over $\omega$ are represented by the choice of $p(\omega)$ and govern the coverage of the Pareto front. 
To examine the effect of $p(\omega)$, which controls the coverage of the Pareto front, we set it to $\text{Dirichlet}(\alpha)$ and vary $\alpha \in \{0.1, 1, 10 \}$.  This results in $\omega$ being sampled from different regions of $\Delta^d$. Specifically, $\alpha=1$ corresponds to a uniform distribution over $\Delta^d$, $\alpha>1$ is skewed towards the center of $\Delta^d$ whereas $\alpha<1$ is skewed towards the corners of $\Delta^d$. In \Cref{fig:ablation_prior} we observe that $\alpha = 1$ results in the best performance. Despite the skewed distribution with $\alpha=0.1$ and $\alpha=10$, we still achieve performance close to that of $\alpha=1$ indicating that MOGFN-PC is able to interpolate to preferences not sampled during training. %Note that diversity is not affected significantly by $p(\omega)$.

\paragraph{Choice of scalarization $R(x|\omega)$.}
The set of $R(x|\omega)$ for different $\omega$ specifies the family of MOO sub-problems and thus has a critical impact on the Pareto performance. \Cref{fig:ablation_scalar_fn} illustrates results for the Weighted Sum (WS), Weighted-log-sum (WL) and Weighted Tchebycheff  (WT) scalarizations. Note that we do not compare the Top-K Reward as different scalarizations cannot be compared directly. WS scalarization results in the best performance. We suspect the poor performance of WT and WL are in part also due to the less smooth reward landscapes they induce.

\paragraph{Impact of $\beta$.}
During training, $\beta$ controls the concentration of the reward density around modes of the distribution. For large values of $\beta$, the reward density around the modes become more peaky and vice-versa. In \Cref{fig:ablation_beta} we present the results obtained by varying $\beta \in \{16, 32, 48, 96\}$. As $\beta$ increases, MOGFN-PC is incentivized to generate samples closer to the modes of $R(x|\omega)$, resulting in better Pareto performance. However, with high $\beta$ values, the reward density is concentrated close to the modes and there is a negative impact on the diversity of the candidates. High values of $\beta$ also make the optimization harder, resulting in poorer performance. As such $\beta$ is a task specific parameter which can be tuned to identify the best trade-off between Pareto performance and diversity but also affects training efficiency.

%% file: tex/appendix.tex
\appendix
\newpage

\section{Algorithms}
We summarize the algorithms for MOGFN-PC and MOGFN-AL here.  
\label{app:algorithms}

\begin{algorithm}
\begin{algorithmic}

\STATE \textbf{Input}:
\STATE $p(\omega)$: Distribution for sampling preferences\;
\STATE $\beta$: Reward Exponent\;
\STATE $\delta$: Mixing Coefficient for uniform actions in sampling policy\;
\STATE $N$: Number of training steps\;
\STATE \textbf{Initialize}:\\
$\left(P_F(s'|s, \omega), P_B(s|s', \omega), \log Z(\omega)\right)$: Conditional GFlowNet with parameters $\theta$\;
\FOR{$i = 1\enspace \text{to} \enspace N$}

    \STATE Sample preference $\omega \sim p(\omega)$\;
    \STATE Sample trajectory $\tau$ following policy $\hat{\pi} =  (1-\delta) P_F \enspace + \delta\text{Uniform}$ \; 
    \STATE Compute reward $R(x|\omega)^\beta$ for generated samples and corresponding loss $\mathcal{L}(\tau, \omega;\theta)$ as in Equation~\ref{eq:mogfn_tb}\;
    \STATE Update parameters $\theta$ with gradients from the loss, $\nabla_\theta \mathcal{L}(\tau, \omega)$\;
\ENDFOR
\end{algorithmic}

\caption{Training preference-conditional GFlowNets}
\label{algo:mogfn-pc}
\end{algorithm}

\begin{algorithm}
\begin{algorithmic}

\STATE \textbf{Input}:
\STATE $\mathbf{R} = \{R_1,\dots,R_d\}$: Oracles to evaluate candidates $x$ and return true objectives $(R_1(x),\dots,R_d(x))$ \;
\STATE $D_0 = \{(x_i, y_i)\}$: Initial dataset with $y_i = \mathbf{R}(x_i)$\;
\STATE $\hat{f}$: Probabilistic surrogate model to model posterior over $\mathbf{R}$ given a dataset $\mathcal{D}$\;%Trainable learner providing functions $\hat{f}.\mu$ and $\hat{f}.\sigma$ with $\mu(x)$ estimating $E[Y|x]$ and $\sigma(x)$ estimating the epistemic uncertainty around $\mu(x)$\;
\STATE $a(x|\hat{f})$: Acquisition function computing a scalar utility for $x$ given $\hat{f}$\;%taking $x$ and $\hat{f}$. Uses $\hat{f}.\mu$ and $\hat{f}.\sigma$ functions to return a reward function $R$\;
\STATE $\pi_\theta$: Learnable GFlowNet policy\;% $a(x|\hat{f})$\;
% $m$: Number of mutations to apply to non-dominated candidates during batch generation\;
\STATE $b$: Size of candidate batch to be generated\;
\STATE $N$: Number of active learning rounds\;
% $K$: Number of top-scoring candidates to keep for $TopK$ evaluation\;
% \textbf{Result}:\\
% $TopK(D_N)$ elements $(x,y) \in D_n$ with highest values of $y$\;
\STATE \textbf{Initialize}:\\
$\hat{f}, \pi_\theta$\;
\FOR{$i = 1\enspace \text{to} \enspace M$}

   \STATE Fit $\hat{f}$ on dataset $D_{i-1}$\;
   \STATE Extract the set of non-dominated candidates $\hat{\mathcal{P}}_{i-1}$ from $D_{i-1}$\;
   \STATE Train $\pi_\theta$ with to generate mutations for $x\in \hat{\mathcal{P}}_i$ using $a(- | \hat{f})$ as the reward\;
   \STATE Generate batch $\mathcal{B} = \{x'_{1,m_i},\ldots,x'_{b,m_b}\}$ by sampling $x'_i$ from $\hat{\mathcal{P}}_{i-1}$ and applying to it mutations $m_i$ sampled from $\pi_\theta$\;
   \STATE Evaluate batch $\mathcal{B}$ with $\mathbf{R}$ to generate $\hat{D}_i = \{(x_1,\mathbf{R}(x_1)),\ldots,(x_b,\mathbf{R}(x_b))\}$\;
   \STATE Update dataset $D_i = \hat{D}_i \cup D_{i-1}$\;
\ENDFOR
\STATE \textbf{Result}:\\
Approximate Pareto set $\hat{\mathcal{P}}_N$
\end{algorithmic}

\caption{Training MOGFN-AL}
\label{algo:mogfn-al}
\end{algorithm}

% \section{Scalarization}
% \label{app:wls_scalarization}
% Scalarization is a popular approach for tackling multi-objective optimization problems. MOGFN-PC can build upon any scalarization approach. We consider three choices. Weighted-sum (WS) scalarization has been widely used in literature. WS finds candidates on the convex hull of the Pareto front~\citep{ehrgott2005multicriteria}. Under the assumption that the Pareto front is convex, every Pareto optimal solution is a solution to a weighted sum problem and the solution to every weighted sum problem is Pareto optimal. Weigthed Tchebycheff (WT), proposed by~\citet{choo1983proper} is an alternative designed for non-convex Pareto fronts. Any Pareto optimal solution can be found by solving the weighted Tchebycheff problem with appropriate weights, and the solutions for any weights correspond to a \textit{weakly} Pareto optimal solution of the original problem~\citep{pardalos2017non}. \citet{lin2021pareto} deomstrated through their empirical results that WT can be used with neural network based policies. The third scheme we consider, Weighted-log-sum (WL) has not been considered in prior work. We hypothesized that in some practical scenarios, we might want to ensure that all objectives are optimized, since, for instance, in WS the scalarized reward can be dominated by a single reward. WL, which considers the weigthed sum in $\log$ space can potentially help with this drawback. However, as discussed in \Cref{sec:analysis}, in practice WL can be hard to optimize, and lead to poor performance.

\section{Additional Analysis}
\label{app:additional_ablations}

% \begin{table}[h]
%     \centering
%     \setlength{\tabcolsep}{2pt}
%     \renewcommand{\arraystretch}{1.5}
%     \caption{\textbf{Fragment-based molecule generation:} Analysing the impact of $\alpha$, $\beta$ and $R(x |\omega)$ on the performace of MOGFN-PC}
% \resizebox{\textwidth}{!}{
%     \begin{tabular}{c c c c |  c c c c|  c c c}
%     \toprule
%   Metrics& \multicolumn{3}{c}{Effect of $p(\omega)$} & \multicolumn{4}{c}{Effect of $\beta$} & \multicolumn{3}{c}{Choice of $R(x \mid \omega)$} \\
%     \hline
%     & Dir(0.1) & Dir(1) & Dir(10) & 16 & 32 & 48 & 96 & WS & WL & WT \\
%     \hline
%     Reward ($\uparrow$) & \g{0.57}{0.04}& \highlight{\g{0.89}{0.05}} & \g{0.58}{0.03}&  \g{0.44}{0.02}&  \g{0.51}{0.008}&  \g{0.55}{0.008} &\highlight{\g{0.89}{0.05}}& -- & -- & --\\
%     Diversity ($\uparrow$) &  \highlight{\g{0.79}{0.01}}&  \g{0.75}{0.01} & \g{0.75}{0.09}&  \g{0.86}{0.006}&  \highlight{\g{0.86}{0.001}} &  \g{0.85}{0.002}& \g{0.75}{0.01} & \g{0.75}{0.01} & \highlight{\g{0.82}{0.016}} & \g{0.10}{0.002}\\
%     Hypervolume ($\uparrow$)& \g{0.67}{0.08}&  \highlight{\g{0.90}{0.01}} &  \g{0.82}{0.12}& \g{0.59}{0.06} &  \g{0.55}{0.001}&  \g{0.60}{0.06}& \highlight{\g{0.90}{0.01}} &\highlight{\g{0.90}{0.01}}&  \g{0.55}{0.017}&  \g{0.71}{0.10}\\
%     $R_2$ ($\downarrow$) & \g{2.57}{0.43}&  \highlight{\g{1.86}{0.08}} & \g{1.93}{1.12}&  \g{5.76}{0.30} &  \g{4.46}{0.28}& \g{3.64}{0.19}& \highlight{\g{1.86}{0.08}} &\highlight{\g{1.86}{0.08}}& \g{6.92}{0.18} & \g{11.51}{1.79}\\
%     \hline
%     \end{tabular}
%     }
%     \label{tab:ablation_mols}
% \end{table}

\begin{figure}[h!]
  	\centering
    \begin{subfigure}[b]{0.45\textwidth}
        \includegraphics[width=1\textwidth]{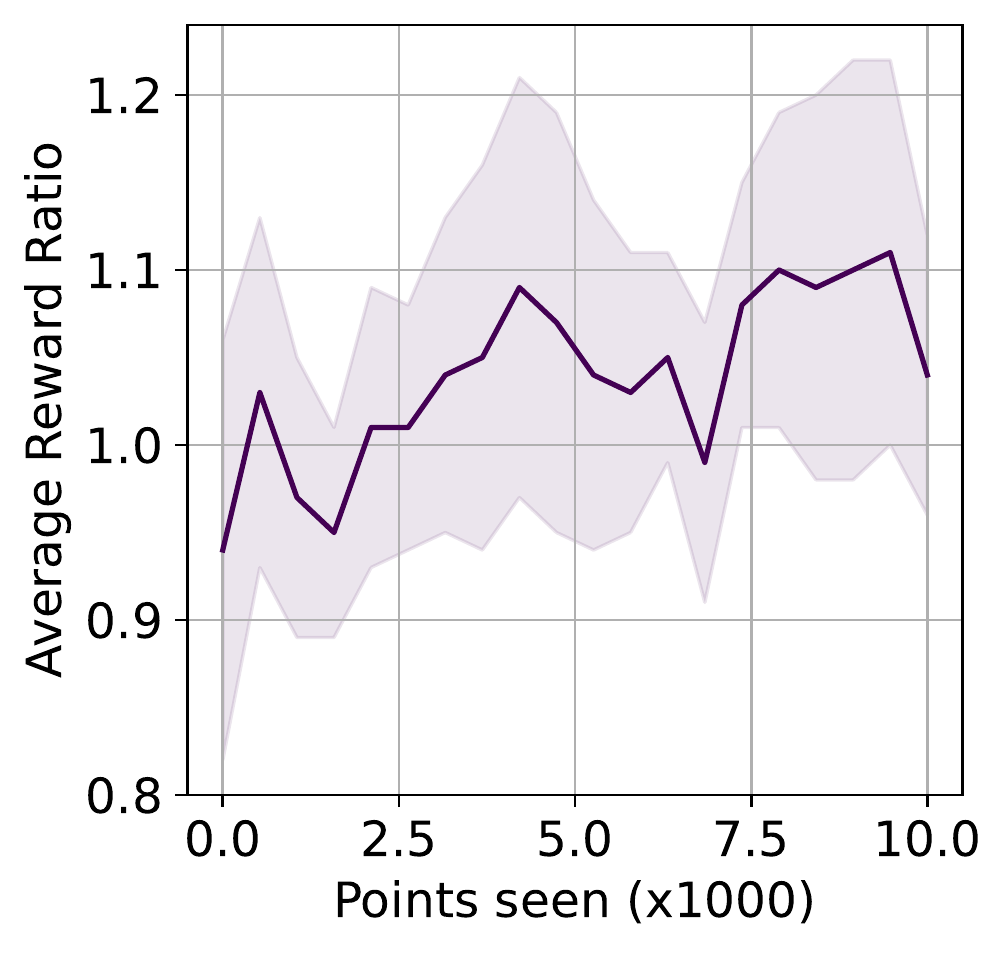}
        \caption{3  Bigrams task}
        \label{fig:sgfn_comp_str}
    \end{subfigure}
    \begin{subfigure}[b]{0.45\textwidth}
        \includegraphics[width=1\textwidth]{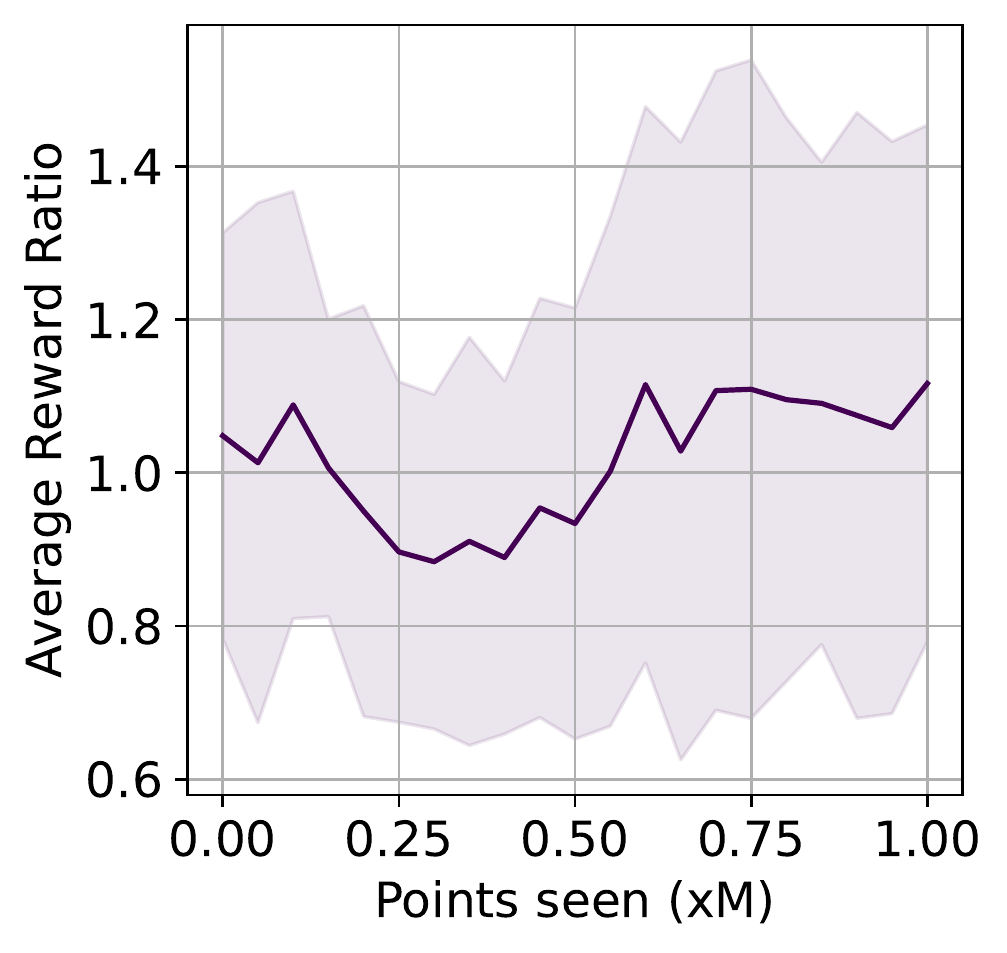}
        \caption{Fragment-based Molecule Generation Task}
        \label{fig:sgfn_comp_frag}
    \end{subfigure}
    
    \caption{We plot the ratio of rewards $R_{mo} / R_{so}$ for candidates generated with MOGFN-PC ($R_mo$) and single-objective GFlowNets($R_{so})$ for a set of preferences in the (a) 3 Bigrams and (b) Fragment-based molecule generation tasks. We observe that MOGFN-PC matches and occasionally surpasses single objective GFlowNets}
    \label{fig:sgfn_comp}
\end{figure}

\paragraph{Can MOGFN-PC match Single Objective GFNs?}
To evaluate how well MOGFN-PC models the family of rewards $R(x|\omega)$, we consider a comparison with single objective GFlowNets. More specifically, we first sample a set of 10 preferences ${\omega_1, \dots, \omega_{10}}$, and train a standard single objective GFlowNet using the weighted sum scalar reward for each preference. We then generate $N=128$ candidates from each GFlowNet, throughout training, and compute the mean reward for the top 10 candidates for each preference. We average this top 10 reward across $\{\omega_1,\dots,\omega_{10}\}$, and call it $R_{so}$. We then train MOGFN-PC, and apply the sample procedure with the preferences $\{\omega_1,\dots,\omega_{10}\}$, and call the resulting mean of top 10 rewards $R_{mo}$. We plot the value of the ratio $R_{mo} / R_{so}$ in \Cref{fig:sgfn_comp}. We observe that the ratio stays close to 1, indicating that MOGFN-PC can indeed model the entire family of rewards simultaneously at least as fast as a single objective GFlowNet could. 

\paragraph{Effect of Model Capacity and Architecture}
Finally we look at the effect of model size in training MOGFN-PC. As MOGFN-PC models a conditional distribution, an entire family of functions as we've described before, we expect capacity to play a crucial role since the amount of information to be learned is higher than for a single-objective GFN. We increase model size in the 3 Bigrams task to see that effect, and see in \Cref{tab:ablation_model_size} that larger models do help with performance--although the performance plateaus after a point. We suspect that in order to fully utilize the model capacity we might need better training objectives.

\begin{table}[ht]
    \centering
    \setlength{\tabcolsep}{2pt}
    \renewcommand{\arraystretch}{1.5}
    \caption{Analysing the impact of model size on the performance of MOGFN-PC. Each architecture choice for the policy is denoted as A-B-C where A is number of layers, B is the number of hidden units in each layer, and  C is the number of attention heads.}
% \resizebox{\textwidth}{!}{
    \begin{tabular}{c c c c c}
    \toprule
   Metrics& \multicolumn{4}{c}{Effect of model size} \\
    \hline
    & 3-64-8 & 3-256-8 & 4-64-8 & 4-256-8 \\
    \hline
    Reward ($\uparrow$) & \g{0.44}{0.01} &  \g{0.47}{0.00} & \g{0.49}{0.03} &  \highlight{\g{0.51}{0.01}} \\
    Diversity ($\uparrow$) &  \highlight{\g{19.79}{0.08}} &  \g{17.13}{0.38} & \g{17.53}{0.15}&  \g{16.12}{0.04} \\
    Hypervolume ($\uparrow$)& \g{0.22}{0.017} & \g{0.255}{0.008}  & \highlight{\g{0.262}{0.003}} & \highlight{\g{0.270}{0.011}} \\
    $R_2$ ($\downarrow$) & \g{9.97}{0.45} &  \g{9.22}{0.25} & \highlight{\g{8.95}{0.05}} &  \highlight{\g{8.91}{0.12}} \\
    \hline
    \end{tabular}
    % }
    \label{tab:ablation_model_size}
\end{table}

\section{Metrics}
\label{app:metrics}
In this section we discuss the various metrics that we used to report the results in \Cref{sec:emp_results}.

\begin{enumerate}
    \item \textbf{Generational Distance Plus (GD +)} \citep{10.1007/978-3-319-15892-1_8}: This metric measures the euclidean distance between the solutions of the Pareto approximation and the true Pareto front by taking the dominance relation into account. To calculate \textbf{GD+} we require the knowledge of the true Pareto front and hence we only report this metric for Hypergrid experiments (\Cref{sec:grid})
    \item \textbf{Hypervolume (HV) Indicator} \citep{1688440}: This is a standard metric reported in \ac{MOO} works which measures the volume in the objective space with respect to a reference point spanned by a set of non-dominated solutions in Pareto front approximation.  
    \item $\boldsymbol{R}_2$\textbf{ Indicator} \citep{hansen1994evaluating}:
    $R_2$ provides a monotonic metric comparing two Pareto front approximations using a set of uniform reference vectors and a utopian point $z^*$ representing the ideal solution of the \ac{MOO}. 
    % Generally, $R_2$ metric calculations are performed with $z^*$ equal to the origin and all objectives transformed to a minimization setting thereby ensuring monotonicity of the metric. Here we are instead in a (reward) maximization setting and so the utopian point considered is the vector of maximal values for each reward (typically 1 for normalized rewards).%This holds true for our experiments as well.
    
    This metric provides a monotonic reference to compare different Pareto front approximations relative to a utopian point. Specifically, we define a set of uniform reference vectors $\lambda \in \Lambda$ that cover the space of the \ac{MOO} and then calculate:$
 R_2(\Gamma, \Lambda, z^*) = \frac{1}{\lvert\Lambda\rvert} \sum_{\substack{\lambda \in \Lambda}} \min_{\substack{\gamma \in \Gamma}} \bigg\{ \max_{\substack{i \in {1, \ldots, k}} } \{\lambda_i \lvert z_i^* - \gamma_i \rvert \}    \bigg\}$
    where $\gamma \in \Gamma$ corresponds to the set of solutions in a given Pareto front approximations and $z^*$ is the utopian point corresponding to the ideal solution of the \ac{MOO}. Generally, $R_2$ metric calculations are performed with $z^*$ equal to the origin and all objectives transformed to a minimization setting, which serves to preserve the monotonic nature of the metric. This holds true for our experiments as well.

    %This metric measures the difference between utility of two Pareto front approximations. In our case we use weighted tchebycheff scalarizing function to measure the utility. 
    \item \textbf{Top-K Reward} This metric was originally used in \citep{bengio2021flow}, which we extend  for our multi-objective setting. For MOGFN-PC, we sample $N$ candidates per test preference and then pick the top-$k$ candidates ($k < N$) with highest scalarized rewards and calculate the mean. We repeat this for all test preferences enumerated from the simplex and report the average top-k reward score.  
    \item \textbf{Top-K Diversity} This metric was also originally used in \citep{bengio2021flow}, which we again extend for our multi-objective setting. We use this metric to quantify the notion of diversity of the generated candidates. Given a distance metric $d(x, y)$ between candidates $x$ and $y$ we calculate the diversity of candidates as those who have $d(x, y)$ greater than a threshold $\epsilon$. For MOGFN-PC, we sample $N$ candidates per test preference and then pick the top-k candidates based on the diversity scores and take the mean. We repeat this for all test preferences sampled from simplex and report the average top-k diversity score. We use the edit distance for sequences, and $1$ minus the Tanimoto similarity for molecules.
    
\end{enumerate}
\section{Additional Experimental Details}

\subsection{Hyper-Grid}
\label{sec:app-hypergrid}
\begin{figure}[ht]
    \centering
    \includegraphics[width=\columnwidth]{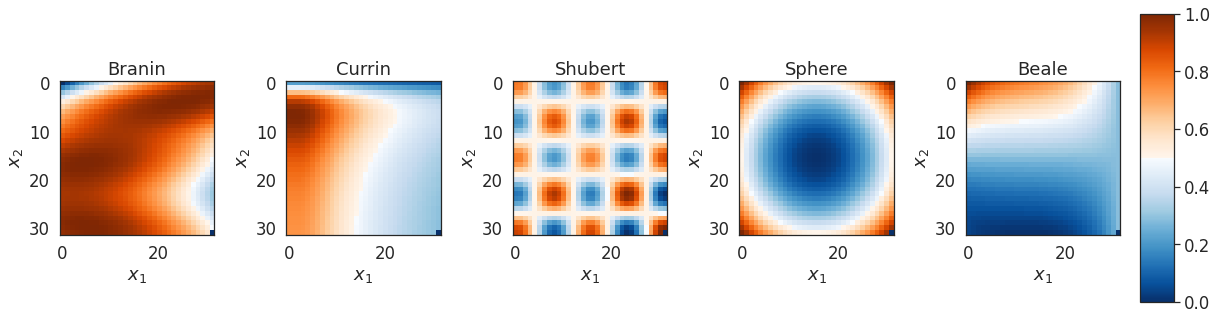}
    \caption{\textbf{Reward Functions} Different reward function considered for HyperGrid experiments presented in \Cref{sec:grid}.  Here the grid dimension is $H=32$}.
    \label{fig:hypergrid-rewards}
\end{figure}

Here we elaborate on the Hyper-Grid experimental setup which we discussed in \Cref{sec:grid}. 
Consider an $n$-dimensional hypercube gridworld where each cell in the grid corresponds to a state. The agent starts at the top left coordinate marked as $(0, 0, \dots)$ and is allowed to move only towards the right, down, or stop. When the agent performs the \textit{stop} action, the trajectory terminates and the agent receives a non-zero reward. In this work, we consider the following reward functions - $\texttt{brannin(x), currin(x), sphere(x), shubert(x), beale(x)}$.
In \Cref{fig:hypergrid-rewards}, we show the heatmap for each reward function. Note that we normalize all the reward functions between 0 and 1.  

% For all our experiments we consider the following rewards: $\boldsymbol{R(x)} = \texttt{[brannin(x), currin(x), sphere(x), shubert(x), beale(x)]}$. 

\begin{figure}[ht]
     \centering
     \begin{subfigure}[b]{0.275\textwidth}
         \centering
         \includegraphics[width=0.75\textwidth]{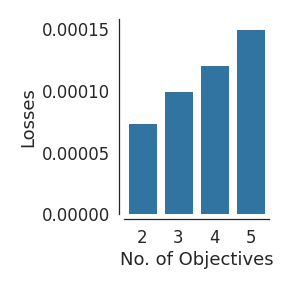}
        %  \caption{Quantitative Results}
         \label{fig:grid_quantitative_1}
     \end{subfigure}
     \begin{subfigure}[b]{0.275\textwidth}
         \centering
         \includegraphics[width=0.75\textwidth]{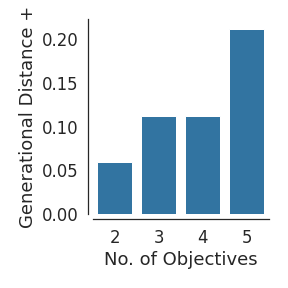}
        %  \caption{Quantitative Results}
         \label{fig:grid_quantitative_2}
     \end{subfigure}
        \caption{\textit{(Left)} Average test loss between the MOGFN-PC distribution and the true distribution for increasing number of objectives. \textit{(Right)} GD $+$ metrics of MOGFN-PC across objectives. } 
        \label{fig:grid_results_app}
\end{figure}

\begin{figure}[h]
    \centering
    \includegraphics[width=\textwidth]{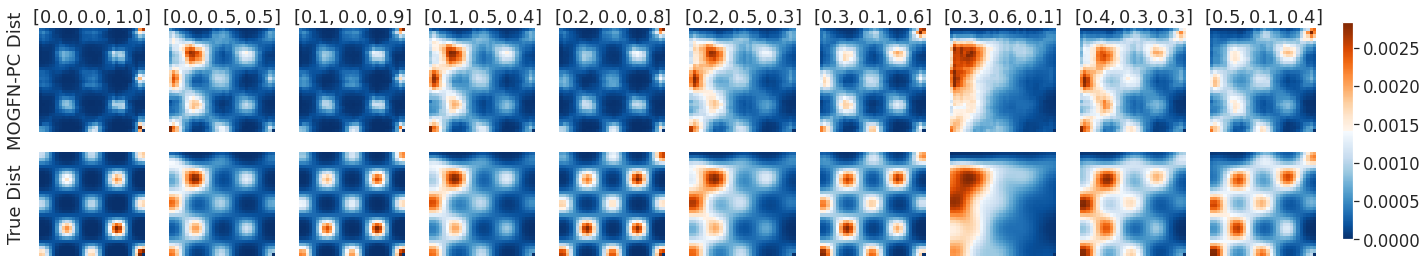}
    \caption{Extended Qualitative Visualizations for Hypergrid epxeriments}
    \label{fig:grid_ext_qualitative}
\end{figure}

\textbf{Additional Results} 
To verify the efficacy of MOGFNs across different objectives sizes, we perform some additional experiments and measure the $L_1$ loss and the $GD+$ metric. In \Cref{fig:grid_results_app}, we can see that as the reward dimension increases, the loss and $GD+$ increases. This is expected because the number of rewards is indicative of the difficulty of the problem. 
We also present extended qualitative visualizations across more preferences in \Cref{fig:grid_ext_qualitative}. 

\textbf{Model Details and Hyperparameters}
For MOGFN-PC policies we use an MLP with two hidden layers each consisting of 64 units. We use \texttt{LeakyReLU} as our activation function as in \cite{bengio2021flow}. All models are trained with $\texttt{learning rate=0.01}$ with the Adam optimizer \cite{DBLP:journals/corr/KingmaB14} and $\texttt{batch size=128}$. We sample preferences $\omega$ from $\text{Dirichlet}(\alpha)$ where $\alpha=1.5$. We try two encoding techniques for encoding preferences - 1) Vanilla encoding where we just use the raw values of the preference vectors and 2) Thermometer encoding~\citep{buckman2018thermometer}. In our experiments we have not observed significant difference in performance difference.  
% \subsubsection{M}

\subsection{N-grams Task}
\label{sec:app-string}
\textbf{Task Details} The task is to generate sequences of some maximum length $L$, which we set to $36$ for the experiments in~\Cref{sec:string}. We consider a vocabulary (actions) of size $21$, with $20$ characters \texttt{["A", "R", "N", "D", "C", "E", "Q", "G", "H", "I", "L", "K", "M", "F", "P", "S", "T", "W", "Y", "V"]} and a special token to indicate the end of sequence. The rewards $\{R_i\}_{i=1}^d$ are defined by the number of occurrences of a given set of n-grams in a sequence $x$. For instance, consider \texttt{["AB", "BA"]} as the n-grams. The rewards for a sequence $x = \texttt{ABABC}$ would be $[2, 1]$. We consider two choices of n-grams: (a) Unigrams: the number of occurrences of a set of unigrams induces conflicting objectives since we cannot increase the number of occurrences of a monogram without replacing another in a string of a particular length, (b) Bigrams: given common characters within the bigrams, the occurrences of multiple bigrams can be increased simultaneously within a string of a fixed length. We also consider different sizes for the set of n-grams considered, i.e. different number of objectives. This allows us to evaluate the behaviour of MOGFN-PC on a variety of objective spaces. We summarize the specific objectives used in our experiments in~\Cref{tab:string_tasks}. We normalize the rewards to $[0, 1]$ in our experiments.
\begin{table}[ht]
\centering
\caption{Objectives considered for the N-grams task}
\begin{tabular}{ll}
\hline
Objectives                                     & n-grams                                                   \\\midrule
2 Unigrams & \texttt{["A", "C"]}                 \\
2 Bigrams       & \texttt{["AC", "CV"]}               \\
3 Unigrams & \texttt{["A", "C", "V"]}           \\
3 Bigrams       & \texttt{["AC", "CV", "VA"]}        \\
4 Unigrams & \texttt{["A", "C", "V", "W"]}     \\
4 Bigrams & \texttt{["AC", "CV", "VA", "AW"]} \\
\hline

\end{tabular}

\label{tab:string_tasks}
\end{table}

\textbf{Model Details and Hyperparameters}
We build upon the implementation from~\citet{stanton2022lambo} for the task: \url{https://github.com/samuelstanton/lambo}. For the string generation task, the backward policy $P_B$ is trivial (as there is only one parent for each node $s\in \mathcal{S}$), so we only have to parameterize $P_F$ and $\log Z$. As $P_F(-|s,\omega)$ is a conditional policy, we use a Conditional Transformer encoder as the architecture. This consists of a Transformer encoder~\citep{vaswani2017attention} with 3 hidden layers of dimension $64$ and $8$ attention heads to embed the current state (string generated so far) $s$. We have an MLP which embeds the preferences $\omega$ which are encoded using thermometer encoding with $50$ bins. The embeddings of the state and preferences are concatenated and passed to a final MLP which generates a categorical distribution over the actions (vocabulary token). We use the same architecture for the baselines using a conditional policy -- MOReinforce and MOSoftQL. For Envelope-MOQ, which does not condition on the preferences, we use a standard Transformer-encoder with a similar architecture. We present the hyperparameters we used in \Cref{tab:string_hparam}. Each method is trained for 10,000 iterations with a minibatch size of 128. For the baselines we adopt the official implementations released by the authors for MOReinforce -- \url{https://github.com/Xi-L/PMOCO} and Envelope-MOQ -- \url{https://github.com/RunzheYang/MORL}.
% \subsection{String}

\begin{table}[ht]
\centering
\caption{Hyperparameters for N-grams Task}
\begin{tabular}{ll}
\hline
\textbf{Hyperparameter}      & \textbf{Values} \\ \hline
Learning Rate ($P_F$)        & \{0.01, 0.05, 0.001, 0.005, 0.0001\}\\
Learning Rate ($Z$)          & \{0.01, 0.05, 0.001\}\\
Reward Exponent: $\beta$     & \{16, 32, 48\}\\
Uniform Policy Mix: $\delta$ & \{0.01, 0.05, 0.1\}\\ \hline
\end{tabular}
\label{tab:string_hparam}
\end{table}

\textbf{Additional Results}
\begin{table*}[t!]
% 	\begin{minipage}{0.7\linewidth}
    
		\centering
		\caption{ \textbf{N-Grams Task:} Diversity and Pareto performance of various algorithms on for the 3 Bigrams and 3 Unigrams tasks with MOGFN-PC achieving superior Pareto performance.}
\resizebox{0.95\textwidth}{!}{
    \begin{tabular}{c c c c c c c c c}
    \toprule
  Algorithm  & \multicolumn{4}{c}{3 Bigrams} & \multicolumn{4}{c}{3 Unigrams}  \\
    \midrule
        & Reward ($\uparrow$) & Diversity ($\uparrow$) & HV ($\uparrow$) & $R_2$ ($\downarrow$) & Reward ($\uparrow$) & Diversity ($\uparrow$) & HV ($\uparrow$) & $R_2$ ($\downarrow$) \\
    \hline
    Envelope-MOQ & \g{0.05}{0.04} & \g{0}{0} &  \g{0.012}{0.013} & \g{19.66}{0.66} & \g{0.08}{0.015} & \g{0}{0} & \g{0.023}{0.011} & \g{21.18}{0.72} \\
    MOReinforce & \g{0.12}{0.02} & \g{0}{0} & \g{0.015}{0.021} & \g{20.32}{0.93}& \g{0.03}{0.001} & \g{0}{0} & \g{0.036}{0.009} & \g{21.04}{0.51}\\
    % MOSoftQL & \g{0.0}{0.0} & \g{0.0}{0.0} & \g{0.0}{0.0} & \g{0.0}{0.0}& \g{0.0}{0.0} & \g{0.0}{0.0} & \g{0.0}{0.0} & \g{0.0}{0.0}\\
    MOSoftQL & \g{0.28}{0.03} & \highlight{\g{21.09}{0.65}} & \g{0.093}{0.025} & \g{15.79}{0.23}& \g{0.36}{0.01} & \highlight{\g{23.131}{0.6736}} & \g{0.105}{0.014} & \g{12.80}{0.26}\\
    MOGFN-PC & \highlight{\g{0.44}{0.01}} & \g{19.79}{0.08} & \highlight{\g{0.220}{0.017}} & \highlight{\g{9.97}{0.45}} & \highlight{\g{0.38}{0.00}} & \g{22.71}{0.24} & \highlight{\g{0.121}{0.015}} & \highlight{\g{11.39}{0.17}}\\
    \hline
    
    \end{tabular}
    }
	
	\label{tab:stringregex3}
\vspace{-6mm}
\end{table*}

We present some additional results for the n-grams task. First, \cref{tab:stringregex3} summarizes the numerical results for the experiments in \autoref{sec:string}. Further, We consider different number of objectives $d \in \{2, 4\}$ in \Cref{tab:synth_string_2obj} and \Cref{tab:synth_string_4obj} respectively. As with the experiments in~\Cref{sec:string} we observe that MOGFN-PC outperforms the baselines in Pareto performance while achieving high diversity scores. In \Cref{tab:synth_string_short}, we consider the case of shorter sequences $L=24$. MOGFN-PC continues to provide significant improvements over the baselines. There are two trends we can observe considering the N-grams task holistically: 
\begin{enumerate}
    \item As the sequence size increases the advantage of MOGFN-PC becomes more significant.
    \item The advantage of MOGFN-PC increases with the number of objectives.
\end{enumerate}

\begin{table}[ht]
    \centering
    \setlength{\tabcolsep}{2pt}

    \renewcommand{\arraystretch}{1.5}
    \caption{\textbf{N-grams Task}. 2 Objectives}
\resizebox{0.95\textwidth}{!}{
    \begin{tabular}{c c c c c c c c c}
    \toprule
  Algorithm  & \multicolumn{4}{c}{2 Bigrams} & \multicolumn{4}{c}{2 Unigrams}  \\
    \midrule
        & Reward ($\uparrow$) & Diversity ($\uparrow$) & HV ($\uparrow$) & $R_2$ ($\downarrow$) & Reward ($\uparrow$) & Diversity ($\uparrow$) & HV ($\uparrow$) & $R_2$ ($\downarrow$) \\
    
    \hline
    Envelope-MOQ & \g{0.05}{0.001} & \g{0}{0} &  \g{0.0}{0.0} & \g{7.74}{0.42} & \g{0.09}{0.02} & \g{0}{0} & \g{0.014}{0.001} & \g{5.73}{0.09} \\
    MOReinforce & \g{0.12}{0.01} & \g{0}{0} & \g{0.151}{0.023} & \g{0.031}{}& \g{0.43}{0.04} & \g{0}{0} & \g{0.222}{0.013} & \g{2.54}{0.06}\\
    MOSoftQL & \g{0.37}{0.03} & \g{19.40}{0.91} & \g{0.247}{0.031} & \g{2.92}{0.39}& \g{0.46}{0.02} & \highlight{\g{22.05}{0.04}} & \g{0.253}{0.003} & \g{2.54}{0.02}\\
    MOGFN-TB & \highlight{\g{0.51}{0.04}} & \highlight{\g{20.65}{0.58}} & \highlight{\g{0.321}{0.011}} & \highlight{\g{2.31}{0.04}} & \highlight{\g{0.48}{0.01}} & \highlight{\g{22.15}{0.22}} & \highlight{\g{0.267}{0.007}} & \highlight{\g{2.24}{0.03}}\\
    \hline
    \end{tabular}
    }
    
    \label{tab:synth_string_2obj}
\end{table}

\begin{table}[ht]
    \centering
    \setlength{\tabcolsep}{2pt}

    \renewcommand{\arraystretch}{1.5}
    \caption{\textbf{N-grams Task}. 4 Objectives}
\resizebox{0.95\textwidth}{!}{
    \begin{tabular}{c c c c c c c c c}
    \toprule
  Algorithm  & \multicolumn{4}{c}{4 Bigrams} & \multicolumn{4}{c}{4 Unigrams}  \\
    \midrule
        & Reward ($\uparrow$) & Diversity ($\uparrow$) & HV ($\uparrow$) & $R_2$ ($\downarrow$) & Reward ($\uparrow$) & Diversity ($\uparrow$) & HV ($\uparrow$) & $R_2$ ($\downarrow$) \\
    
    \hline
    Envelope-MOQ & \g{0}{0} & \g{0}{0} &  \g{0}{0} & \g{85.23}{2.78} & \g{0}{0} & \g{0}{0} & \g{0}{0} & \g{80.36}{3.16} \\
    MOReinforce & \g{0.01}{0.00} & \g{0}{0} & \g{0.001}{0.001} & \g{60.42}{1.52}& \g{0.00}{0.00} & \g{0}{0} & \g{0}{0} & \g{79.12}{4.21}\\
    MOSoftQL & \g{0.12}{0.04} & \highlight{\g{24.32}{1.21}} & \g{0.013}{0.001} & \g{39.31}{1.35}& \g{0.22}{0.02} & \highlight{\g{24.18}{1.43}} & \g{0.019}{0.005} & \g{31.46}{2.32}\\
    MOGFN-TB & \highlight{\g{0.23}{0.02}} & \g{20.31}{0.43} & \highlight{\g{0.055}{0.017}} & \highlight{\g{24.42}{1.44}} & \highlight{\g{0.33}{0.01}} & \g{23.24}{0.23} & \highlight{\g{0.063}{0.032}} & \highlight{\g{23.31}{2.03}}\\
    \hline
    \end{tabular}
    }
    
    \label{tab:synth_string_4obj}
\end{table}

\begin{table}[ht]
    \centering
    \caption{\textbf{N-grams Task}. Shorter Sequences}
    \setlength{\tabcolsep}{2pt}

    \renewcommand{\arraystretch}{1.5}
\resizebox{0.95\textwidth}{!}{
    \begin{tabular}{c c c c c c c c c}
    \toprule
  Algorithm  & \multicolumn{4}{c}{3 Bigrams} & \multicolumn{4}{c}{3 Unigrams}  \\
    \midrule
        & Reward ($\uparrow$) & Diversity ($\uparrow$) & HV ($\uparrow$) & $R_2$ ($\downarrow$) & Reward ($\uparrow$) & Diversity ($\uparrow$) & HV ($\uparrow$) & $R_2$ ($\downarrow$) \\
    
    \hline
    Envelope-MOQ & \g{0.07}{0.01} & \g{0}{0} &  \g{0.027}{0.010} & \g{16.21}{0.48} & \g{0.08}{0.02} & \g{0}{0} & \g{0.031}{0.015} & \g{20.13}{0.41} \\
    MOReinforce & \g{0.18}{0.01} & \g{0}{0} & \g{0.053}{0.031} & \g{13.35}{0.82}& \g{0.07}{0.02} & \g{0}{0} & \g{0.041}{0.009} & \g{19.25}{0.41}\\
    % MOSoftQL & \g{0.0}{0.0} & \g{0.0}{0.0} & \g{0.0}{0.0} & \g{0.0}{0.0}& \g{0.0}{0.0} & \g{0.0}{0.0} & \g{0.0}{0.0} & \g{0.0}{0.0}\\
    MOSoftQL & \g{0.31}{0.02} & \highlight{\g{20.12}{0.51}} & \g{0.143}{0.019} & \g{12.79}{0.41}& \highlight{\g{0.38}{0.02}} & \g{21.13}{0.35} & \g{0.109}{0.011} & \g{12.12}{0.24}\\
    MOGFN-PC & \highlight{\g{0.45}{0.02}} & \g{19.62}{0.04} & \highlight{\g{0.225}{0.009}} & \highlight{\g{9.82}{0.23}} & \highlight{\g{0.39}{0.01}} & \highlight{\g{21.94}{0.21}} & \highlight{\g{0.125}{0.015}} & \highlight{\g{10.91}{0.14}}\\
    \hline
    \end{tabular}
    }
    
    \label{tab:synth_string_short}
\end{table}

\FloatBarrier

\subsection{QM9}
\label{sec:app-qm9}
\textbf{Reward Details} As mentioned in \Cref{sec:qm9}, we consider four reward functions for our experiments. The first reward function is the HUMO-LUMO gap, for which we rely on the predictions of a pretrained MXMNet~\citep{zhang2020molecular} model trained on the QM9 dataset \citep{ramakrishnan2014quantum}. The second reward is the standard Synthetic Accessibility score which we calculate using the RDKit library~\citep{rdkit}, to get the reward we compute $(10-\texttt{SA})/9$. The third reward function is molecular weight target. Here we first calculate the molecular weight of a molecule using RDKit, and then construct a reward function of the form $e^{-(\texttt{molWt} - 105)^2/150}$ which is maximized at 105. Our final reward function is a logP target, $e^{-(\texttt{logP} - 2.5)^2/2}$, which is again calculated with RDKit and is maximized at 2.5.

\textbf{Model Details and Hyperparameters}
 We sample new preferences for every episode from a $Dirichlet(\alpha)$, and encode the desired sampling temperature using a thermometer encoding \citep{buckman2018thermometer}. We use a graph neural network based on a graph transformer architecture~\citep{yun2019graph}. We transform this conditional encoding to an embedding using an MLP. The embedding is then fed to the GNN as a virtual node, as well as concatenated with the node embeddings in the graph. The model's action space is to add a new node to the graph, a new bond, or set node or bond properties (like making a bond a double bond). It also has a \texttt{stop} action. For more details please refer to the code provided in the supplementary material. We summarize the hyperparameters used in \Cref{tab:hp_qm9}. 
\begin{table}[ht!]
\centering
\begin{tabular}{ll}
\hline
\textbf{Hyperparameter}      & \textbf{Value} \\ \hline
Learning Rate ($P_F$)        &    0.0005        \\
Learning Rate ($Z$)          &  0.0005      \\
Reward Exponent: $\beta$     &  32             \\
Batch Size:                  &  64           \\
%Sampling $\tau$              &  0           \\
Number of Embeddings         &  64            \\
Uniform Policy Mix: $\delta$ &  0.001               \\ 
Number of layers             &  4             \\ \hline
\hline
\end{tabular}
\caption{Hyperparameters for QM9 Task}
\label{tab:hp_qm9}

\end{table}

\FloatBarrier

\subsection{Fragments}
\label{sec:app-fragments}
\textbf{More Details}
As mentioned in \Cref{sec:frags}, we consider four reward functions for our experiments. The first reward function is a proxy trained on molecules docked with AutodockVina~\citep{trott2010autodock} for the sEH target; we use the weights provided by \citet{bengio2021flow}. We also use synthetic accessibility, as for QM9, and a weight target \emph{region} (instead of the specific target weight used for QM9), \texttt{((300 - molwt) / 700 + 1).clip(0, 1)} which favors molecules with a weight of under 300. Our final reward function is QED which is again calculated with RDKit.

\textbf{Model Details and Hyperparameters}
We again use a graph neural network based on a graph transformer architecture~\citep{yun2019graph}. The experimental protocol is similar to QM9 experiments discussed in \Cref{sec:app-qm9}. We additionally sample from a lagged model whose parameters are updated as $\theta' = \tau \theta' + (1-\tau)\theta$. The model's action space is to add a new node, by choosing from a list of fragments and an attachment point on the current molecular graph. We list all hyperparameters used in \Cref{tab:hp_frag}. 
\begin{table}[ht!]
\centering
\begin{tabular}{ll}
\hline
\textbf{Hyperparameter}      & \textbf{Value} \\ \hline
Learning Rate ($P_F$)        &   0.0005         \\
Learning Rate ($Z$)          &   0.0005         \\
Reward Exponent: $\beta$     &   96            \\
Batch Size:                  &   256           \\
Sampling model $\tau$              &  0.95           \\
Number of Embeddings         &  128            \\
Number of layers             &  6              \\ \hline
\end{tabular}
\caption{Hyperparameters for Fragments}
\label{tab:hp_frag}

\end{table}

\begin{figure}[ht!]
    \centering
    \includegraphics[width=0.75\columnwidth]{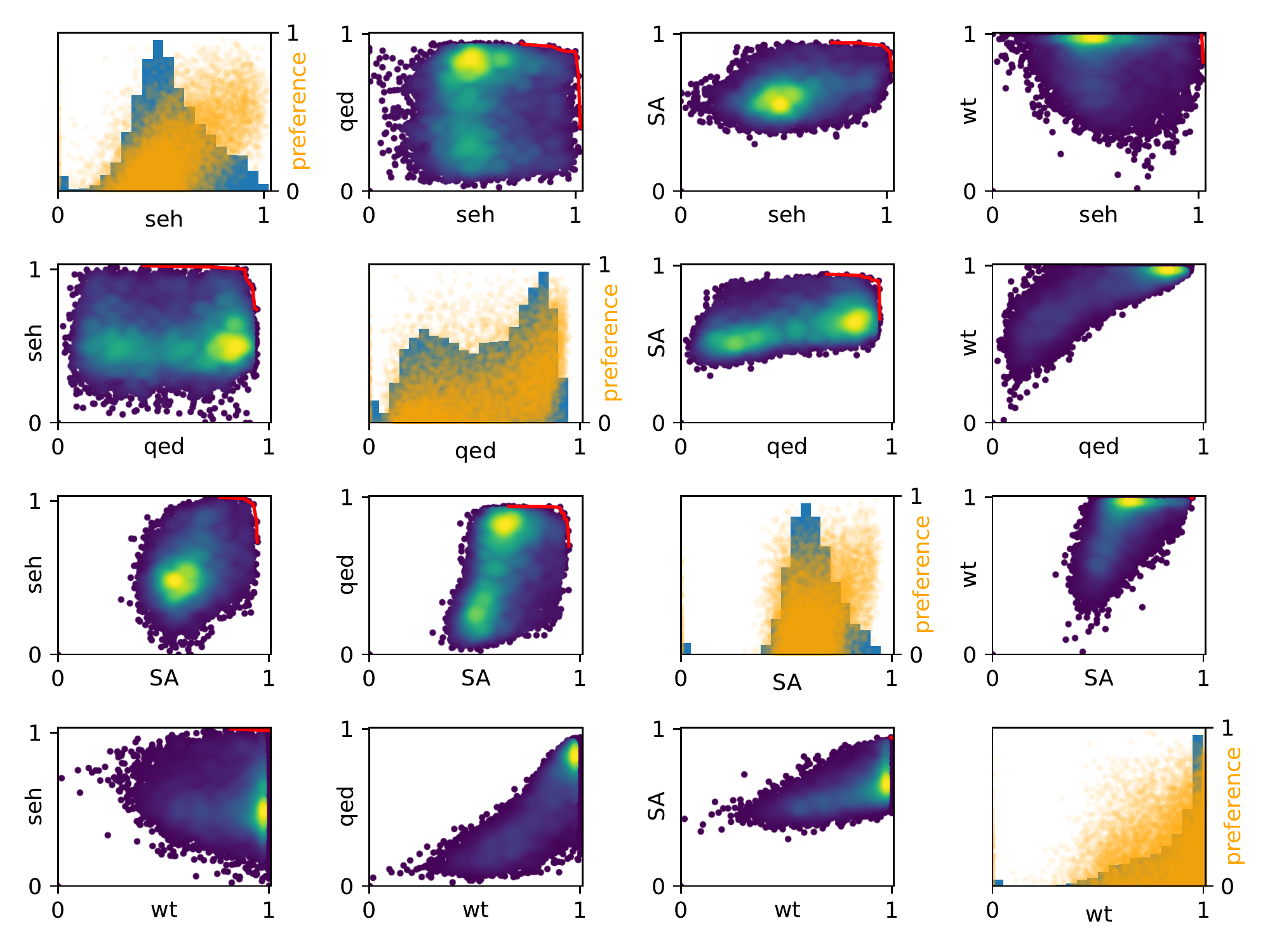}
    \caption{Fragment-based molecule generation: See \Cref{sec:app-fragments}.}
    \label{fig:seh_frag_moo_4x4}
\end{figure}

\textbf{Additional Results}
We also present in \Cref{fig:seh_frag_moo_4x4} a view of the reward distribution produced by MOGFN-PC. Generally, the model is able to find good near-Pareto-optimal samples, but is also able to spend a lot of time exploring. The figure also shows that the model is able to respect the preference conditioning, and remains capable of generating a diverse distribution rather than a single point.

In the off-diagonal plots of \Cref{fig:seh_frag_moo_4x4}, we show pairwise scatter plots for each objective pair; the Pareto front is depicted with a red line; each point corresponds to a molecule generated by the model as it explores the state space; color is density (linear viridis palette). The diagonal plots show two overlaid informations: a blue histogram for each objective, and an orange scatter plot showing the relationship between preference conditioning and generated molecules. The effect of this conditioning is particularly visible for \texttt{seh} (top left) and \texttt{wt} (bottom right). As the preference for the sEH binding reward gets closer to 1, the generated molecules’ reward for sEH gets closer to 1 as well. Indeed, the expected shape for such a scatter plot is a triangular-ish shape: when the preference $\omega_i$ for reward $R_i$ is close to 1, the model is expected to generate objects with a high reward for $R_i$; as the preference $\omega_i$ gets further away from 1, the model can generate anything, including objects with a high $R_i$--that is, unless there is a trade off between objectives, in which case in cannot; this is the case for the \texttt{seh} objective, but not for the \texttt{wt} objective, which has a more triangular shape.

\subsection{DNA Sequence Design}
\label{sec:app-dna}
% \textbf{Additional Results}
\paragraph{Task Details}
The set of building blocks here consists of the bases\texttt{["A", "C", "T", "G"]} in addition to a special end of sequence token. In order to compute the free energy and number of base with the software NUPACK \citep{zadeh2011nupack}, we used 310 K as the temperature. The inverse of the length $L$ objective was calculated as $\frac{30}{L}$, as 30 was the minimum length for sampled sequences. The rewards are normalized to $[0, 1]$ for our experiments.

\paragraph{Model Details and Hyperparameters}
We use the same implementation as the N-grams task, detailed in \Cref{sec:app-string}. Here we consider a 4-layer Transformer architecture, with 256 units per layer and $16$ attention head instead. We detail the most relevant hyperparameters \Cref{tab:dna_hp}.

\begin{table}[ht!]
\centering
\caption{Hyperparameters tuned for DNA-Aptamers Task.}
\begin{tabular}{ll}
\hline
\textbf{Hyperparameter}      & \textbf{Values} \\ \hline
Learning Rate ($P_F$)        & \{0.001, 0.0001, 0.00001, 0.000001\}\\
Learning Rate ($Z$)          & 0.001           \\
Reward Exponent: $\beta$     & \{40, 60, 80\} \\
Batch Size:                  & 16           \\
Training iterations:         & 10,000           \\
Dirichlet $\alpha$           & \{0.1, 1.0, 1.5\} \\ \hline
\end{tabular}
\label{tab:dna_hp}
\end{table}

\paragraph{Discussion of Results}
Contrary to the other tasks on which we evaluated MOGFN-PC, for the generation of DNA aptamer sequences, our proposed model did not match the best baseline, multi-objective reinforcement learning~\citep{lin2021pareto}, in terms of Pareto performance. Nonetheless, it is worth delving into the details in order to better understand the different solutions found by the two methods. First, as indicated in \cref{sec:emp_results}, despite the better Pareto performance, the best sequences generated by the RL method have extremely low diversity (0.62), compared to MOGFN, which generates optimal sequences with diversity of 19.6 or higher. As a matter of fact, MOReinforce mostly samples sequences with the well-known pattern \texttt{GCGC...} for all possible lengths. Sequences with this pattern have indeed low (negative) energy and many number of pairs, but they offer little new insights and poor diversity if the model is not able to generate sequences with other distinct patterns. On the contrary, GFlowNets are able to generate sequences with patterns other than repeating the pair of bases \texttt{G} and \texttt{C}. Interestingly, we observed that GFlowNets were able to generate sequences with even lower energy than the best sequences generated by MOReinforce by inserting bases \texttt{A} and \texttt{T} into chains of \texttt{GCGC...}. Finally, we observed that one reason why MOGFN does not match the Pareto performance of MOReinforce is because for short lengths (one of the objectives) the energy and number of pairs are not successfully optimised. Nonetheless, the optimisation of energy and number of pairs is very good for the longest sequences. Given these observations, we conjecture that there is room for improving the set of hyperparameters or certain aspects of the algorithm.

\paragraph{Additional Results}
In order to better understand the impact of the main hyperparameters of MOGFN-PC in the Pareto performance and diversity of the optimal candidates, we train multiple instances by sweeping over several values of the hyperparameters, as indicated in \Cref{tab:dna_hp}. We present the results in \Cref{tab:dna-ablation}. One key observation is that there seems to be a tradeoff between the Pareto performance and the diversity of the Top-K sequences. Nonetheless, even the models with the lowest diversity are able to generate much more diverse sequences than MOReinforce. Furthermore, we also observe $\alpha<1$ as the parameter of the Dirichlet distribution to sample the weight preferences, as well as higher $\beta$ (reward exponent), both yield better metrics of Pareto performance but slightly worse diversity. In the case of $\beta$, this observation is consistent with the results of the analysis in the Bigrams task (\Cref{fig:ablation}), but with Bigrams, best performance was obtained with $\alpha=1$. This is indicative of a degree of dependence on the task and the nature of the objectives.

\begin{table}[ht!]
    \centering
    \setlength{\tabcolsep}{2pt}
    \renewcommand{\arraystretch}{1.5}
    \caption{Analysis of the impact of $\alpha$, $\beta$ and the learning rate on the performance of MOGFN-PC for DNA sequence design. We observe a trade-off between the Top-K diversity and the Pareto performance.}
\resizebox{\textwidth}{!}{
    \begin{tabular}{c c c c |  c c c | c c c c}
    \toprule
    Metrics & \multicolumn{3}{c}{Effect of $p(\omega)$} & \multicolumn{3}{c}{Effect of $\beta$} & \multicolumn{4}{c}{Effect of the learning rate} \\
    \hline
                             & Dir($\alpha=0.1$)           & Dir($\alpha=1$) & Dir($\alpha=1.5$)           & 40                          & 60              & 80                          & $10^{-5}$        & $10^{-4}$        & $10^{-3}$                 & $10^{-2}$ \\
    \hline
    Reward ($\uparrow$)      & \highlight{\g{0.687}{0.01}} & \g{0.652}{0.01} & \g{0.639}{0.01}             & \g{0.506}{0.01}             & \g{0.560}{0.01} & \highlight{\g{0.652}{0.01}} & \g{0.587}{0.01} & \g{0.652}{0.01} & \highlight{\g{0.654}{0.03}} & \g{0.604}{0.01}             \\
    Diversity ($\uparrow$)   & \g{17.65}{0.37}             & \g{19.58}{0.15} & \highlight{\g{20.18}{0.58}} & \highlight{\g{28.49}{0.32}} & \g{24.93}{0.19} & \g{19.58}{0.15}             & \g{21.92}{0.59} & \g{19.58}{0.15} & \g{19.51}{1.14}             & \highlight{\g{23.16}{0.18}} \\
    Hypervolume ($\uparrow$) & \highlight{\g{0.506}{0.01}} & \g{0.467}{0.02} & \g{0.440}{0.01}             & \g{0.277}{0.03}             & \g{0.363}{0.03} & \highlight{\g{0.467}{0.02}} & \g{0.333}{0.01} & \g{0.467}{0.02} & \highlight{\g{0.496}{0.01}} & \g{0.336}{0.01}             \\
    $R_2$ ($\downarrow$)     & \highlight{\g{2.462}{0.05}} & \g{2.576}{0.08} & \g{2.688}{0.02}             & \g{4.225}{0.34}             & \g{2.905}{0.18} & \highlight{\g{2.576}{0.08}} & \g{3.855}{0.31} & \g{2.576}{0.01} & \highlight{\g{2.488}{0.03}} & \g{3.422}{0.07}             \\
    \hline
    \end{tabular}
    }
    \label{tab:dna-ablation}
\end{table}

\subsection{Active Learning}
\label{sec:app-activelearning}
\textbf{Task Details}
We consider the Proxy RFP task from~\citet{stanton2022lambo}, an in silico benchmark task designed to simulate searching for improved red fluorescent protein
(RFP) variants~\citep{dance2021hunt}. The objectives considered are stability (-dG or negative change in Gibbs free energy) and solvent-accessible surface area
(SASA)~\citep{shrake1973environment} in simulation, computed using the FoldX suite~\citep{schymkowitz2005foldx} and BioPython~\citep{cock2009biopython}. We use the dataset introduced in~\citet{stanton2022lambo} as the initial pool of candidates $\mathcal{D}_0$ with $|\mathcal{D}_0|=512$.

\textbf{Method Details and Hyperparameters}
Our implementation builds upon the publicly released code from~\citep{stanton2022lambo}: \url{https://github.com/samuelstanton/lambo}. 
We follow the exact experimental setup used in~\cite{stanton2022lambo}. The surrogate model $\hat{f}$ consists of an encoder with 1D convolutions (masking positions corresponding to padding tokens). We used $3$ standard pre-activation residual blocks with two convolution layers, layer norm, and swish activations, with a kernel size of 5, 64 intermediate channels and 16 latent channels. A multi-task GP with an ICM kernel is defined in the latent space of this encoder, which outputs the predictions for each objective. We also use the training tricks detailed in~\citet{stanton2022lambo} for the surrogate model. The hyperparameters, taken from~\citet{stanton2022lambo} are shown in \Cref{tab:surrogate_params}. The acquisiton function used is NEHVI~\citep{daulton2021parallel} defined as 

\begin{equation}
    \alpha(\{x_j\}_{j=1}^i) = \frac{1}{N}\sum_{t=1}^N\text{HVI}(\{\tilde{f}_t(x_j)\}_{j=1}^{i-1}|\mathcal{P}_t) + \frac{1}{N}\sum_{t=1}^N\text{HVI}(\tilde{f}_t(x_j)|\mathcal{P}_t\cup \{\tilde{f}_t(x_j)\}_{j=1}^{i-1})
\end{equation}

where $\tilde{f}_t, t={1,\dots N}$ are independent draws from the surrogate model (which is a posterior over functions), and $\mathcal{P}_t$ denotes the Pareto frontier in the current dataset $\mathcal{D}$ under $\tilde{f}_t$. 
\begin{table}[ht!]
\centering
\caption{Hyperparameters for training the surrogate model $\hat{f}$}
\begin{tabular}{ll}
\hline
Hyperparameter                         & Value        \\ \hline
Shared enc. depth (\# residual blocks) & 3            \\
Disc. enc. depth (\# residual blocks)  & 1            \\
Decoder depth (\# residual blocks)     & 3            \\
Conv. kernel width (\# tokens)         & 5            \\
\# conv. channels                      & 64           \\
Latent dimension                       & 16           \\
GP likelihood variance init            & 0.25         \\
GP lengthscale prior                   & N(0.7, 0.01) \\
\# inducing points (SVGP head)         & 64           \\
DAE corruption ratio (training)        & 0.125        \\
DAE learning rate (MTGP head)          & 5.00E-03     \\
DAE learning rate (SVGP head)          & 1.00E-03     \\
DAE weight decay                       & 1.00E-04     \\
Adam EMA params                        & 0., 1e-2     \\
Early stopping holdout ratio           & 0.1          \\
Early stopping relative tolerance      & 1.00E-03     \\
Early stopping patience (\# epochs)    & 32           \\
Max \# training epochs                 & 256          \\ \hline
\end{tabular}
\label{tab:surrogate_params}
\end{table}

We replace the LaMBO candidate generation with GFlowNets. We generate a set of mutations $m = \{(l_i, v_i)\}$ for a sequences $x$ from the current approximation of the Pareto front $\hat{\mathcal{P}}_i$. Note that, as opposed to the sequence generation experiments, $P_B$ here is not trivial as there are multiple ways (orders) of generating the set. For our experiments, we use a uniform random $P_B$. $P_F$ takes as input the sequence $x$ with the mutations generated so far applied. We use a Transformer encoder with 3 layers, with hidden dimension 64 and 8 attention heads as the architecture for the policy. The policy outputs a distribution over the locations in $x$, $\{1, \dots, |x|\}$, and a distribution over tokens for each location. The vocabulary of actions here is the same as the N-grams task - \texttt{["A", "R", "N", "D", "C", "E", "Q", "G", "H", "I", "L", "K", "M", "F", "P", "S", "T", "W", "Y", "V"]}. The logits of the locations of the mutations generated so far are set to -1000, to prevent generating the same sequence. The acquisition function(NEHVI) value for the mutated sequence is used as the reward. We also use a reward exponent $\beta$. To make optimization easier (as the acquisition function becomes harder to optimize with growing $\beta$), we reduce $\beta$ linearly by a factor $\delta \beta$ at each round. We train the GFlowNet for 750 iterations in each round. \Cref{tab:al_hparam} shows the MOGFN-AL hyperparameters. The active learning batch size is 16, and we run 64 rounds of optimization. Table~\ref{tab:al_hparam} presents the hyperparameters used for MOGFN-AL.

\begin{table}[ht!]
\centering
\caption{Hyperparameters for MOGFN-AL}
\begin{tabular}{ll}
\hline
\textbf{Hyperparameter}      & \textbf{Values} \\ \hline
Learning Rate ($P_F$)        & \{0.01, 0.001, 0.0001\}\\
Learning Rate ($Z$)          & \{0.01, 0.001\}\\
Reward Exponent: $\beta$     &\{16, 24\}\\
Uniform Policy Mix: $\delta$ &\{0.01, 0.05\}\\
Maximum number of mutations     &\{10, 15, 20\}\\
$\delta \beta$     &\{0.5, 1, 2\}\\\hline
\end{tabular}
\label{tab:al_hparam}
\end{table}